\documentclass{article}

\usepackage[preprint]{arxiv_styles/neurips_arxiv_style}

\usepackage{natbib} 
\usepackage{mathtools} 
\usepackage{booktabs} 
\usepackage{tikz} 

\usepackage{amsmath,amsfonts,bm}

\def\eqref#1{Equation~\ref{#1}}

\def\1{\bm{1}}

\DeclareMathAlphabet{\mathsfit}{\encodingdefault}{\sfdefault}{m}{sl}
\SetMathAlphabet{\mathsfit}{bold}{\encodingdefault}{\sfdefault}{bx}{n}

\newcommand{\E}{\mathbb{E}}

\DeclareMathOperator{\Tr}{Tr}

\usepackage{graphicx, subcaption}
\usepackage{hyperref}
\usepackage{booktabs}
\usepackage{amsthm}
\usepackage{amsfonts}
\usepackage{url}
\usepackage[title]{appendix}

\usepackage{amsmath}
\usepackage{amsthm}
\usepackage{mathtools}
\usepackage{amsfonts,dsfont,eucal}

\newcommand\extrafootertext[1]{%
    \bgroup
    \renewcommand\thefootnote{\fnsymbol{footnote}}%
    \renewcommand\thempfootnote{\fnsymbol{mpfootnote}}%
    \footnotetext[0]{#1}%
    \egroup
}

\newtheorem{property}{Property}

\newcommand{\DiME}{\operatorname{DiME}_{\alpha}}

\title{DiME: Maximizing Mutual Information by a \\ Difference of Matrix-Based Entropies}

\author{
  Oscar Skean$^1$ 
   \And
  Jhoan Keider Hoyos Osorio$^1$ 
  \AND
  Austin J. Brockmeier$^2$ 
    \And
  Luis Gonzalo Sanchez Giraldo$^1$
}

\begin{document}
\maketitle

\begin{abstract}
We introduce an information-theoretic quantity with similar properties to mutual information that can be estimated from data without making explicit assumptions on the underlying distribution. This quantity is based on a recently proposed matrix-based entropy that uses the eigenvalues of a normalized Gram matrix to compute an estimate of the eigenvalues of an uncentered covariance operator in a reproducing kernel Hilbert space. We show that a difference of matrix-based entropies (DiME) is well suited for problems involving the maximization of mutual information between random variables. While many methods for such tasks can lead to trivial solutions, DiME naturally penalizes such outcomes. We compare DiME to several baseline estimators of mutual information on a toy Gaussian dataset. We provide examples of use cases for DiME, such as latent factor disentanglement and a multiview representation learning problem where DiME is used to learn a shared representation among views with high mutual information.
\end{abstract}

\extrafootertext{
    $^1$ \normalfont University of Kentucky \quad\quad
    $^2$ \normalfont University of Delaware
}
    
\extrafootertext{Our source code is available at \href{https://github.com/uk-cliplab/DiME}{https://github.com/uk-cliplab/DiME}}
\extrafootertext{Correspondence to oscar.skean@uky.edu}

\section{Introduction}\label{sec:intro}
Quantifying the dependence between variables is a fundamental problem in science. Mutual information (MI) is one descriptor of dependence and is widely utilized in information theory to measure the dependence between two random variables $ X$ and $ Y$.
MI has been employed in numerous fields, such as statistics \citep{zhang2022normal}, manufacturing \citep{zhao2022adversarial}, health \citep{huntington2021cytokine}, and  machine learning \citep{sanchez2020learning, tschannen2019mutual, hjelm2018learning, belghazi2018mine}.

Mutual information measures the amount of information shared by $X$ and $Y$ in terms of the reduction in uncertainty about $X$ by observing $Y$ or vice versa. Let $\mathcal{X}$ and $\mathcal{Y}$ be the respective ranges (alphabets) of $ X$ and  $Y$ and $\mathbb{P}_{X,Y}$ their joint distribution defined on $\mathcal{X} \times \mathcal{Y}$. The MI between $X$ and $Y$ is defined as $   I(X, Y) = H(X) - H(X\vert Y) = H(Y) - H(Y\vert X)$,
where 
$H(\cdot)$ is the entropy of a given random variable, $H(X \vert Y)$ is the conditional entropy of $X$ given $Y$, and vice versa. Computing the MI for continuous multidimensional random variables is not trivial as it requires knowledge of the distribution $\mathbb{P}_{X,Y}$ or joint probability density function in cases of absolute continuity. A natural way to estimate MI is to estimate the underlying probability density functions from data samples \citep{orlitsky2003always,orlitsky2015competitive} to later approximate the MI; however, estimating probability densities is not data efficient \citep{majdara2022efficient}. Other works, such as the one proposed in \citep{kraskov2004estimating}, are based on entropy estimates from k-nearest neighbor distances, although this method requires making assumptions on the underlying distributions that might not  be true. While some of these approaches work well in an information-theoretic sense, many are inefficient and have difficulty scaling to high-dimensional data. Given the notorious difficulties of measuring MI from high-dimensional datasets, popular alternative methods have emerged for maximizing (or minimizing) a lower (upper) bound on MI \citep{poole2019variational, mcallester2020formal, cheng2020club} that is more tractable and scalable.\par 

Despite the popularity of maximizing MI estimators for learning representations, it has been observed that maximizing very tight bounds on MI can lead to inferior representations \citep{tschannen2019mutual}. Intuitively, the quality of the representations is determined more by the architecture employed than the MI estimator used. Indeed, maximizing MI between the representation of paired views of the same instances does not guarantee lower MI between the representations of unpaired instances. Therefore, we propose a formulation that explicitly captures both the maximization of mutual information between paired instances and the minimization of mutual information between unpaired instances by using an information-theoretic quantity known as matrix-based entropy \citep{sanchezgiraldo2015measures}. The formulation, \textbf{Di}fference of \textbf{M}atrix-based \textbf{E}ntropies (DiME), compares the joint entropy of the kernel-based similarity matrix of two paired random variables (for example, representations of two different views) to the expected joint entropy of the similarity matrix using unpaired views. The latter is obtained by averaging the joint entropy across multiple random permutations of the instances in one view within a batch. Because DiME is derived from matrix-based entropy quantities, it is not technically an estimator of true MI. Instead, DiME shares important properties with MI and provides a computationally tractable surrogate. \par

Our main contributions are:

\begin{itemize}
    \item We introduce DiME and show that it is readily understood in terms of matrix-based entropies, which can easily be implemented, and excels as an objective function for problems seeking to maximize mutual information.
    \item We show that DiME, while not a mutual information estimator, varies proportionally to mutual information, a property not always exhibited by mutual information estimators. 
    \item We demonstrate the effectiveness of DiME in several tasks, namely: learning shared representations between multiple views and disentangling latent factors.
    \item We show that DiME in combination with matrix-based conditional entropies can be used to disentangle latent factors into independent subspaces while not restricting the latent subspace dimensions to be independent.
\end{itemize} \par

\section{Background}
Before introducing DiME, we first provide a description of the matrix-based entropy---the basic building block of DiME. We use matrix-based quantities because they serve as tractable surrogates for information-theoretic quantities.

\subsection{Matrix-Based Entropy}
Let $\mathbf{X} = \{x_i\}_{i=1}^{n}$ be a set of $n$ data points $x \in \mathcal{X}$ sampled from an unknown distribution $\mathbb{P}_X$ defined on $\mathcal{X}$.
Let $\kappa : \mathcal{X} \times \mathcal{X} \mapsto \mathbb{R}_{\geq 0}$ be a positive definite kernel that is normalized such that $\kappa(x, x) = 1$ for all $x \in \mathcal{X}$. We can construct a Gram matrix $\mathbf{K}_{\mathbf{X}}$ consisting of all pairwise evaluations of the points in $\mathbf{X}$. Given $\mathbf{K}_{\mathbf{X}}$, the matrix-based entropy of order $\alpha>0$ is defined as:
\begin{equation}\label{eq:matrix_based_entropy}
S_{\alpha}\left(\mathbf{K}_{\mathbf{X}}\right) = \frac{1}{1-\alpha}\log{\left[\Tr{\left( \left( \frac{1}{n} \mathbf{K}_{\mathbf{X}} \right)^{\alpha}\right) }\right]},
\end{equation}
where $\mathbf{K}^{\alpha}$ is an arbitrary matrix power and $\Tr$ denotes the trace operator which is obtained from the sum of the $\alpha$-power of each of the eigenvalues \citep{bhatia1997}. Essentially, $S_{\alpha}\left(\mathbf{K}_{\mathbf{X}}\right)$ is  Rényi's $\alpha$-order entropy of the eigenvalues of $\frac{1}{n}\mathbf{K}_{\mathbf{X}}$. $S_{\alpha}$ is an information-theoretic quantity that behaves analogously to Rényi's $\alpha$-order entropy, but it can be estimated directly from data without making strong assumptions about the underlying distribution \citep{sanchezgiraldo2015measures, sanchezgiraldo2013iclr}. 



\subsection{Matrix-Based Joint entropy}
Let $\mathcal{X}$ and $\mathcal{Y}$ be two nonempty sets for which there is a joint probability measure space $\left(\mathcal{X} \times \mathcal{Y}, \mathbf{B}_{\mathcal{X} \times \mathcal{Y}}, \mathbb{P}_{X,Y}\right)$ for a set $\left\{(x_i,y_i)\right\}_{i=1}^n$ of $n$ pairs sampled from a joint distribution $\mathbb{P}_{X,Y}$. By defining a kernel $\kappa : \left(\mathcal{X}\times\mathcal{Y}\right) \times \left(\mathcal{X}\times\mathcal{Y}\right) \rightarrow \mathbb{R}$ we can extend \eqref{eq:matrix_based_entropy} to pairs of variables. A choice consistent with kernels on $\mathcal{X}$ and $\mathcal{Y}$ is the product kernel:
\begin{equation}\label{eq:product_kernel}
\kappa_{\mathcal{X} \times \mathcal{Y}}((x,y), (x', y')) = \kappa_{\mathcal{X}}(x, x')\kappa_{\mathcal{Y}}(y, y').
\end{equation} 
This corresponds to the tensor product between all dimensions of the representations of $\mathcal{X}$ and $\mathcal{Y}$. For kernels $\kappa_{\mathcal{X}}$ and $\kappa_{\mathcal{Y}}$, such that $\kappa_{\mathcal{X}}(x,x) = 1$, the product kernel is equivalent to concatenating the dimensions of the Hilbert spaces resulting from taking the $\log$ of the kernel. For instance, if $\mathcal{X} \subset \mathbb{R}^{d_\mathcal{X}}$ and $\mathcal{Y} \subset \mathbb{R}^{d_\mathcal{Y}}$ and we use the Gaussian kernel for both  $\kappa_{\mathcal{X}}$ and $\kappa_{\mathcal{Y}}$, the product kernel is a kernel on $\mathbb{R}^{d_\mathcal{X} + d_\mathcal{Y}}$, which corresponds to direct concatenation of features in the input space.
This concatenation leads to the notion of matrix-based joint entropy, which can be expressed in terms of \eqref{eq:matrix_based_entropy} using the Hadamard product of the kernel matrices, as
$S_{\alpha}(\mathbf{K}_\mathbf{X} \circ \mathbf{K}_\mathbf{Y})$.


\subsection{Matrix-Based Conditional Entropy and Mutual Information}
The $\alpha$-order matrix-based mutual information is defined as
\begin{equation}\label{eq:representation_mutual_information}
I_{\alpha}(\mathbf{K}_\mathbf{X} ; \mathbf{K}_\mathbf{Y}) =  S_{\alpha}(\mathbf{K}_\mathbf{X}) - S_{\alpha}(\mathbf{K}_\mathbf{X} \vert \mathbf{K}_\mathbf{Y}) = S_{\alpha}(\mathbf{K}_\mathbf{X}) +  S_{\alpha}(\mathbf{K}_\mathbf{Y}) - S_{\alpha}(\mathbf{K}_\mathbf{X} \circ \mathbf{K}_\mathbf{Y}), 
\end{equation} 
where the matrix-based conditional entropy is defined as
\begin{equation}\label{eq:represenation_conditional_entroopy}
S_{\alpha}(\mathbf{K}_\mathbf{X} \vert \mathbf{K}_\mathbf{Y}) = S_{\alpha}(\mathbf{K}_\mathbf{X} \circ \mathbf{K}_\mathbf{Y}) - S_{\alpha}(\mathbf{K}_\mathbf{Y}). 
\end{equation}
These quantities are well behaved in the limit $\alpha \rightarrow 1$, where $S_1$ corresponds to von Neumann entropy (Shannon entropy of the eigenvalues) of the trace-normalized matrices. When $\alpha\neq 1$, it has been shown that subadditivity, $S_{\alpha}(\mathbf{K}_\mathbf{X} \circ \mathbf{K}_\mathbf{Y}) \leq S_{\alpha}(\mathbf{K}_\mathbf{X}) +  S_{\alpha}(\mathbf{K}_\mathbf{Y})$, still holds \citep{camilo2019physrev}, yielding $I_{\alpha}(\mathbf{K}_\mathbf{X} ; \mathbf{K}_\mathbf{Y})\ge 0$. But $S_{\alpha}(\mathbf{K}_\mathbf{X} \vert \mathbf{K}_\mathbf{Y}) \leq S_{\alpha}(\mathbf{K}_\mathbf{X})$ is not always true for cases of $\alpha\neq1$ \citep{teixeira2012conditional}. Nevertheless, larger values of $\alpha$ can still be useful to emphasize high-density regions of the data. Here $\alpha=1.01$ is used in all experiments. \par

It must be emphasized that even when $\alpha\rightarrow 1$ the matrix-based quantities are not estimators of Shannon's entropy nor mutual information. For instance, it is possible for the differential entropy of continuous random variables to be negative, whereas matrix-based entropy is always non-negative. Also, if two continuous random variables are equal, such as when $X=Y$ or $X$ and $Y$ are related through an invertible mapping, Shannon's mutual information $I(X, Y)$ is infinite. In contrast, $I_{\alpha}(\mathbf{K}_\mathbf{X} ; \mathbf{K}_\mathbf{X}) \leq  S_{\alpha}(\mathbf{K}_\mathbf{X}) \leq \log n$ with an upper-bound that depends on the sample size, exhibiting properties similar to those of discrete random variables with support equal to the sample size. However, scaling the variables themselves does affect matrix-based entropy, which is a property of continuous random variables, discussed in the following section.

\section{Difference of Matrix-Based Entropies (DiME)}
\label{sect:dime-intro}
The matrix-based mutual information between paired samples from random variables $X$ and $Y$ introduced in \eqref{eq:representation_mutual_information} works well in practice \citep{sanchezgiraldo2015measures, zhang2022icassp}, but it does require proper selection of kernel parameters due the following properties.

For $\alpha > 0$ and non-negative normalized kernels $\mathbf{K}_\mathbf{X}$ and $\mathbf{K}_\mathbf{Y}$, we can readily verify that:
\begin{eqnarray}\label{eq:joint_entropy_as_upper_bound}
S_{\alpha}(\mathbf{K}_\mathbf{X} \circ \mathbf{K}_\mathbf{Y}) & \geq & S_{\alpha}(\mathbf{K}_\mathbf{X}) \\
S_{\alpha}(\mathbf{K}_\mathbf{X} \circ \mathbf{K}_\mathbf{Y}) & \geq & S_{\alpha}(\mathbf{K}_\mathbf{Y}).
\end{eqnarray}
These inequalities lead to $S_{\alpha}(\mathbf{K}_\mathbf{X} \circ \mathbf{K}_\mathbf{X}) \geq  S_{\alpha}(\mathbf{K}_\mathbf{X})$ and the following property.
\begin{property}\label{prop:entropy_exponent} Let $\mathbf{K}$ be a normalized Gram matrix and $\mathbf{K}^{\circ \gamma}$ denote the matrix of entry-wise $\gamma$ power. If $\mathbf{K}$ is an infinitely divisible matrix, that is $\mathbf{K}^{\circ \gamma}$ is positive semidefinite for any non-negative $\gamma$, $S_{\alpha}(\mathbf{K}^{\circ \gamma})$ is a monotonically increasing function of $\gamma$. 
\begin{equation}
\mathrm{S}_{\alpha}(\mathbf{K}^{\circ \gamma_1}) \leq \mathrm{S}_{\alpha}(\mathbf{K}^{\circ \gamma_2}),
\end{equation}
for $0 < \gamma_1 \leq \gamma_2$.
\end{property}

The Gaussian kernel $\kappa(x,x')=e^{\frac{-1}{2\sigma^2}\lVert x-x'\rVert_2^2}$ creates infinitely divisible Gram matrices, and taking the entry-wise exponent of the Gram matrix is equivalent to changing the width $\sigma$ of the kernel or scaling the data $\kappa(x,x')^{\gamma_1}=e^{\frac{-1}{2(\sigma/\sqrt{\gamma_1})^2}\lVert x-x'\rVert_2^2 }=e^{\frac{-1}{2\sigma^2}\lVert \sqrt{\gamma_1}x-\sqrt{\gamma_1}x'\rVert_2^2}$. As in differential entropy,  as we scale the random variable relative to a fixed $\sigma$, we can have larger or smaller matrix-based entropies for the same set of points. Because of property \ref{prop:entropy_exponent}, decreasing the kernel size $\sigma$ leads to a trivial maximization of \eqref{eq:representation_mutual_information} where  $I_{\alpha}(\mathbf{K}_{\mathbf{X}} ; \mathbf{K}_{\mathbf{Y}}) = \log{n}$. On the other hand, a large kernel size makes $I_{\alpha}(\mathbf{K}_{\mathbf{X}} ; \mathbf{K}_{\mathbf{Y}})$ very small and unable to capture dependencies between $X$ and $Y$. 

Inspired by hypothesis testing for independence, where the matrix-based MI between paired samples $\mathbf{K}_{\mathbf{X}}$ and $\mathbf{K}_{\mathbf{Y}}$ is compared to a surrogate for the null distribution by sampling values of matrix-based MI between $\mathbf{K}_{\mathbf{X}}$ and $\bm{\Pi}\mathbf{K}_{\mathbf{Y}}\bm{\Pi}^{T}$, where $\bm{\Pi}$ is a random permutation matrix, we propose using the following difference: 
\begin{equation}\label{eq:difference_of_matrix_entropies}
I_{\alpha}(\mathbf{K}_{\mathbf{X}} ; \mathbf{K}_{\mathbf{Y}}) - \E_{\bm{\Pi}}\left[ I_{\alpha}(\mathbf{K}_{\mathbf{X}} ; \bm{\Pi}\mathbf{K}_{\mathbf{Y}}\bm{\Pi}^{T})\right]
=\E_{\bm{\Pi}}\left[ S_{\alpha}(\mathbf{K}_{\mathbf{X}} \circ \bm{\Pi}\mathbf{K}_{\mathbf{Y}}\bm{\Pi}^{T})\right] - S_{\alpha}(\mathbf{K}_{\mathbf{X}} \circ \mathbf{K}_{\mathbf{Y}}),
\end{equation}
which follows from the fact that the marginal matrix-based entropy is invariant to permutations.

Crucially, unlike \eqref{eq:representation_mutual_information}, the difference of matrix-based entropies \eqref{eq:difference_of_matrix_entropies} does not monotonically increase as the kernel bandwidth goes to zero. Instead, this difference moves from small to large and back to small as we decrease the kernel size from $\sigma=\infty \rightarrow \sigma=0$ (we provide a graphic for this claim in Appendix \ref{appendix:dime-bandwidth-behavior}). In other words, with the difference of matrix-based entropies we can select the kernel parameter for which the matrix-based joint entropy of a set of points drawn from the joint distribution $\mathbb{P}_{X,Y}$ can be \textit{most} distinguished from the entropy of a random permutation surrogate of the product of marginals $\mathbb{P}_{X}\otimes \mathbb{P}_{Y}$. This difference of matrix-based entropies constitutes a lower bound on the matrix-based mutual information \eqref{eq:representation_mutual_information}. If we parameterize our kernel function, such that $\mathbf{K}_{\mathbf{X}}(\bm{\theta}_X)$ and $\mathbf{K}_{\mathbf{Y}}(\bm{\theta}_Y)$ are now functions of the parameter vector $\bm{\theta} = \{\bm{\theta}_X, \bm{\theta}_Y\}$, this yields DiME: 
\begin{multline}
\label{eq:contrastive_representation_mutual_information}
\DiME(\mathbf{X}; \mathbf{Y}; \bm{\theta}) = 
 \E_{\bm{\Pi}}\left[ S_{\alpha}(\mathbf{K}_{\mathbf{X}}(\bm{\theta}_X) \circ \bm{\Pi}\mathbf{K}_{\mathbf{Y}}(\bm{\theta}_Y) \bm{\Pi}^{T})\right] - S_{\alpha}(\mathbf{K}_{\mathbf{X}}(\bm{\theta}_X) \circ \mathbf{K}_{\mathbf{Y}}(\bm{\theta}_Y))
\end{multline}
We can then maximize the difference of matrix-based entropies \eqref{eq:difference_of_matrix_entropies} with respect to $\bm{\theta}$ as measure of dependence and a surrogate for mutual information. In practice, to estimate the expectation over permutations, we compute an average over a finite number of permutations. We have experimentally observed that even a single permutation can work well, but to decrease variance we use five permutations for most experiments.

A key operation to calculate \eqref{eq:matrix_based_entropy} is eigendecomposition. It is well known that eigendecomposition for a square matrix has a time complexity of $O(n^3)$, where $n$ is the matrix size. For a very large $n$ this operation is prohibitive. In our case,  $n$ is the size of a mini-batch which is typically small. We are thus able to compute DiME in a reasonable time without resorting to potentially faster, but less accurate, eigendecomposition approximations.

\section{Related Work}
\subsection{Mutual Information Estimation}
\label{sect:miestimation}
Methods for estimating MI (or bounds on MI) have led to a plethora of works using MI maximization for unsupervised representation learning problems \citep{sordoni2021decomposed,tian2020contrastive,bachman2019learning,hjelm2018learning, oord2018representation}. Many of these works are inspired by the InfoMax principle introduced by \citep{linsker1988self} to learn a representation that maximizes its MI with the input. However, a more tractable approach is to maximize the MI between the representations of two ``views'' of the input which has been shown to be a lower bound on the InfoMax cost function \citep{tschannen2019mutual}. This approach is convenient since the learned representations are typically encoded in a low-dimensional space.\par 
There are several factors necessary for good performance in MI-based representation learning, such as: the way the views are chosen, the MI estimator, and the network architectures employed. We focus on the MI estimator in particular. Among the common estimators are InfoNCE ($I_{\textrm{NCE}}$) \citep{oord2018representation}, Nguyen, Wainwright, and Jordan ($I_{\textrm{NWJ}}$)~\citep{nguyen2010estimating}, Mutual Information Neural Estimation (MINE) \citep{belghazi2018mine}, CLUB \citep{cheng2020club}, and difference-of-entropies (DoE) \citep{mcallester2020formal}. $I_{\textrm{NCE}}$ aims to maximize a lower bound on MI between a pair of random variables by discriminating positive pairs from negative ones \citep{wu2021rethinking}. $I_{\textrm{NWJ}}$ trains a log density ratio estimator to maximize a variational lower bound on the Kullback-Leibler (KL) divergence. A similar bound, denoted as $I_{\textrm{JS}}$ \citep{hjelm2018learning}, does the equivalent although using the Jensen-Shannon divergence instead. Similarly, MINE is based on a dual representation of the KL divergence and relies on a neural network to approximate the MI. CLUB is a variational upper bound of MI that is particularly suited for MI minimization tasks. Finally, DoE estimates MI as a difference of entropies, bounding the entropies by cross-entropy terms. DoE is  neither an upper nor lower bound of MI but exhibits evidence that it is accurate in estimating large MI values. \par
\subsection{Matrix-based Entropy in Representation Learning}
This work is not the first time that matrix-based entropy has been used for representation learning. In \citet{zhang2022icassp}, a supervised information-bottleneck approach is used to learn joint representations for views and their labels. Matrix-based MI is utilized to simultaneously minimize MI between each view and its encodings and to maximize MI between fused encodings and labels. However, the usage of matrix-based MI implies that kernel bandwidth cannot be optimized simultaneously and is instead chosen for each mini-batch based on the mean distance between sample embeddings. Because the kernel bandwidth is calculated at each mini-batch, there are no guarantees on the scale of the network outputs. One advantage of our proposal is solving selecting bandwidth parameters and forcing the output of the networks to specified ranges necessary to avoid trivial solutions.\par

\subsection{Usage of Permutations in Objective}
Our work uses permutations to decouple random variables. This idea has been used before in FactorVAE \citep{kim2018disentangling}, MINE \citep{belghazi2018mine}, and CLUB \citep{cheng2020club}. In those works, permutations are used to efficiently compute a product of marginal distributions/densities. However, the self-regulating property of DiME, which uses permutations to estimate a joint entropy of independent variables, is novel and open for further exploration.






\section{Comparison to Variational Bounds}
\label{sect:variational-bounds}
We apply DiME to a toy problem described in \citet{poole2019variational}. We use their same specifications to construct the dataset. Specifically, we sample $(X, Y)$ from a 20-dimensional Gaussian with zero mean and correlation $\rho$, where $\rho$ increases every 4000 iterations.  At each iteration, a new batch is drawn and MI estimates are computed. With this setup, the true MI can be computed as $I(X; Y) = -\frac{d}{2} \log (1 - \rho ^2) = -10 \log (1 - \rho ^2)$. The choices of $\rho$ correspond to MI values of $\{2, 4, 6, 8, 10\}$.

The purpose of this experiment is to show how changes in the true MI correspond to changes in MI estimators. Specifically, we take an average of estimates from iteration 3900–4000 (directly before MI increases from 2 to 4) and divide every estimation value by this average. This allows one to examine, for instance, if doubling the true MI means that the estimator doubles too. We show these relative values, rather than actual estimated values, because DiME and matrix-based MI are on different scales to true MI. We choose the range of 3900–4000 so that the estimators are well-trained.

We compare DiME to four variational methods for MI estimation introduced in Section \ref{sect:miestimation}: $I_{\textrm{JS}}$, $I_{\textrm{NWJ}}$, CLUB, and $I_{\textrm{NCE}}$. We also compare to matrix-based mutual information (MBMI). For each of the variational methods, we maximize the bound by training an MLP critic with two hidden layers of 256 units (15 units for CLUB), ReLU activations, and an embedding dimensionality of 32. These methods are in contrast to DiME, where we simply (and optionally) optimize two parameters: the kernel bandwidths in \eqref{eq:contrastive_representation_mutual_information} where $\kappa$ is the Gaussian kernel. We initialize all kernel bandwidths to $\sqrt{d}=\sqrt{20} \approx 4.5$ and use Adam \citep{kingma2014adam} optimizer for all estimators. \par

\begin{figure*}[t]
     \centering
     \begin{subfigure}[b]{\textwidth}
         \centering
         \includegraphics[width=\textwidth]{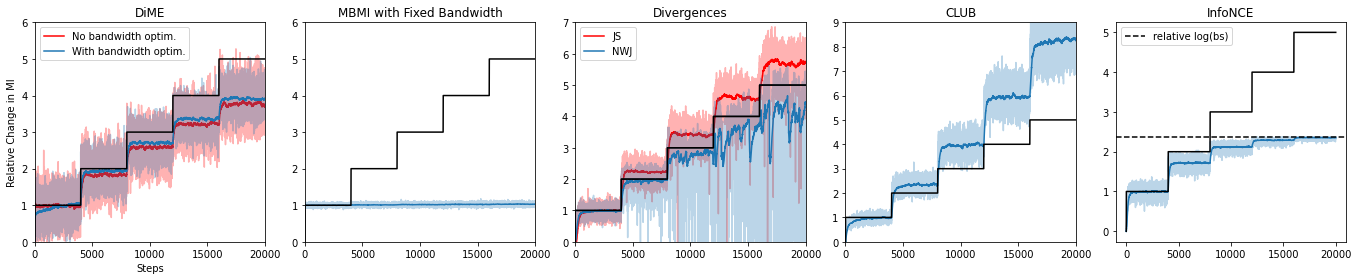}
         \label{fig:variationalbounds_64}
     \end{subfigure}
     \centering
     \begin{subfigure}[b]{\textwidth}
         \centering
         \includegraphics[trim=0 40 0 40, width=\textwidth]{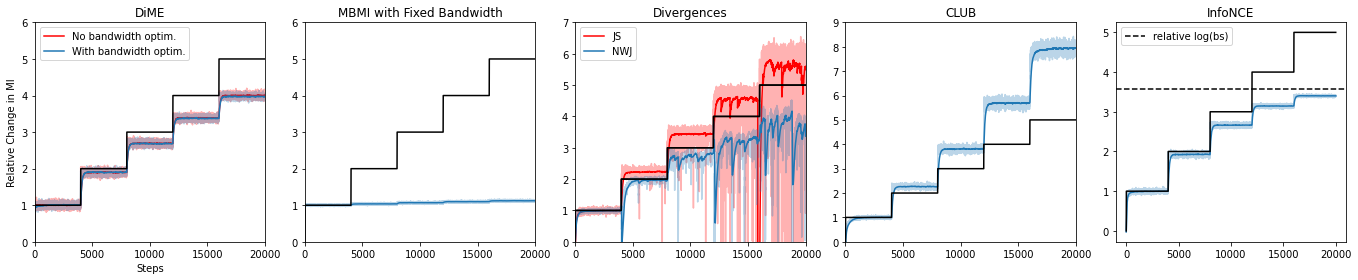}
         \label{fig:variationalbounds_256}
     \end{subfigure}
     \caption{Performance of different estimators of MI for a toy correlated Gaussian dataset with increasing correlation over time. The black staircase line denotes the relative increase of true MI of the dataset. \textbf{(top)} Trained with batch size of 64 and using joint critics for variational estimators. \textbf{(bottom)} Trained with batch size of 1024 and using separable critics for variational estimators. }
    \label{fig:variationalbounds}
\end{figure*}

Results for this experiment are shown in Figure~\ref{fig:variationalbounds}.  The quantity statistics (mean and variance) are computed in a sliding window over the 200 previous iterations. The variance of DiME is less affected by the underlying true MI than $I_{\textrm{JS}}$, $I_{\textrm{NWJ}}$ and CLUB, albeit with a high variance for small batch size. As opposed to $I_{\textrm{NCE}}$, DiME scales well even when MI exceeds the $\log$ of the batch size. Additionally, even without any tuning of kernel bandwidth, DiME still distinguishes between MI levels. However, we observe that training kernel bandwidths using DiME is often helpful. An explanation for why $I_{\textrm{JS}}$ and CLUB exceed the growth of MI is because they underestimate when MI is low. This affects the visualization and makes good estimations at high MI seem artificially large. We provide an alternative plot with the actual estimated values in Appendix \ref{appendix:alternative-value-plot}. Additionally, we discuss how DiME is affected by batch size and dimensionality in the Appendix \ref{appendix:dime-batch-behavior}. 

\section{Experiments}
We present several experiments showcasing potential applications for DiME. The main purpose of these experiments is to highlight the versatility of DiME when combined with other matrix-based information-theoretic quantities. 
As discussed previously, DiME is well suited for tasks involving the maximization of MI between random variables and assigns more value to simpler configurations that exhibit higher dependence. This can be used in contrastive learning and representation learning when trying to maximize dependence between two views of data, e.g.  \citet{wang2015deep, oord2018representation, hjelm2018learning, zbontar2021barlow}. \par
Let $\mathbf{X}^{(1)} = \left\{x_i^{(1)}\right\}_{i=1}^N$ and $\mathbf{X}^{(2)} =  \left\{x_i^{(2)}\right\}_{i=1}^N$ be paired sets of instances from two different views of some latent variable. The goal is to train encoders $f_1\in \mathcal{F}_1$ and $f_2 \in \mathcal{F}_2$ (potentially with shared weights) in order to maximize the MI between the encoded representations by using DiME,
\begin{equation}
    \underset{f_1 \in \mathcal{F}_1, f_2 \in \mathcal{F}_2}{\operatorname{maximize}} \DiME\left(f_1(\mathbf{X}^{(1)}),f_2(\mathbf{X}^{(2)}), \theta \right),
    \label{eq:dime_multiview}
\end{equation}
where $\theta$ is a shared kernel parameter. We apply our objective function in two multiview datasets and assess the results on the downstream tasks of classification and latent factor disentanglement.
\subsection{Multiview MNIST}
\label{subsection:multiview MNIST}
We follow the work of \citet{wang2015deep} in constructing a multiview MNIST dataset with two views. The first view contains digits rotated randomly between $-45$ and $45$ degrees. The second view contains digits with added noise sampled uniformly from $[0, 1]$, clamping to one all values in excess of one. The second view is then shuffled so that paired images are different instances from the same class. With this setup, the only information shared between views is the class label. We thus expect a maximization of dependence between the shared representation between views to carry only the class label information.\par
We encode each view using separate encoders with CNN architectures described in Appendix \ref{appendix:multiview-model}. DiME is optimized between the views as described in~\eqref{eq:dime_multiview}. We use a Gaussian kernel with fixed bandwidth $\theta= \sqrt{D/2}$ for each view. A batch size of 3000 is used.

\begin{figure}[!t]
\centering
\subfloat{{\includegraphics[scale=0.4]{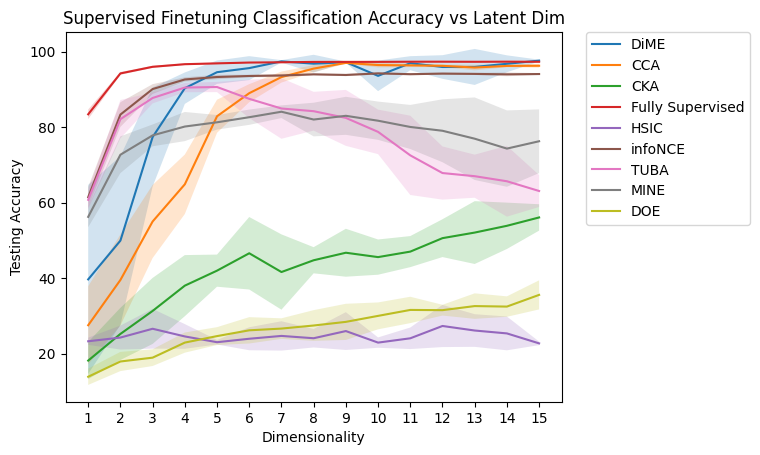}}}%
\subfloat{{\includegraphics[scale=0.40]{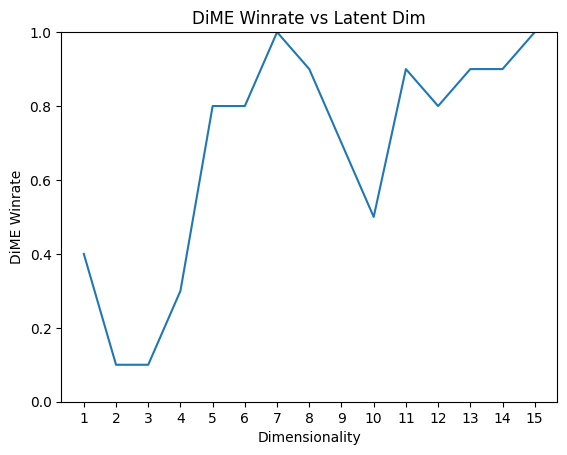}}}%
\caption{\textbf{(left)} Classification accuracy across dimensionality obtained via firstly training encoders with a dependence objective and then training a classifier consisting of one hidden layer after the frozen encoder. The accuracy of equivalently sized networks trained using direct supervision is included. \textbf{(right)} Win rate of DiME against other methods, excluding fully supervised. }
\label{fig:downstream}

\end{figure}

We compare to the following baselines on the task of downstream classification accuracy: DCCA \citep{wang2015deep}, CKA \citep{cortes2012algorithms, kornblith2019similarity}, HSIC \citep{gretton2005measuring}, infoNCE \citep{oord2018representation}, TUBA \citep{poole2019variational}, MINE \citep{belghazi2018mine}, and difference-of-entropies (DoE) \citep{mcallester2020formal}.
We define downstream classification accuracy as training an encoder using a measure of dependence followed by the training of a supervised classifier with one hidden layer on the encoder's output. Note that the encoder parameters are frozen after initial training. While the encoder never has access to labels during training, this approach is not unsupervised since label information is used in the pairings.\par
The downstream classification accuracies for varying latent dimensionalities, averaged over 10 random network initializations, are shown in Figure \ref{fig:downstream}. We also provide DiME's win rate, 
which we define as the proportion of the 10 trials that it is the best performing method (excluding fully supervised training). We observe that DiME wins a majority of the trials for higher latent dimensionalities. DiME also requires fewer latent dimensions to be competitive with the fully supervised training. However, DiME does have a higher variance than other top performing methods for specific dimensionalities.

\subsection{Disentanglement of Latent Factors}
\label{subsection:disentanglement}

So far we have extracted information that is common to two views. It is natural to ask if one can instead extract information that is exclusive to views. With the multiview MNIST dataset, exclusive information to View 1 would be rotations and for View 2 it would be noise. Because pairings are done by class label, exclusive information could also constitute latent factors such as stroke width, boldness, height, and width. 
In order to separate shared and exclusive information, we use the matrix-based conditional entropy defined in \eqref{eq:represenation_conditional_entroopy}. Let $\bm{S}^{(i)}$ and $\bm{E}^{(i)}$ define the shared and exclusive information in view $i\in\{1,2\}$ captured by the encoder $f_{i}$ for data $\bm{X}^{(i)}$. Then $f_i(\bm{X}^{(i)}) = [\bm{S}^{(i)}, \bm{E}^{(i)}]$, where the right side is a concatenation of dimensions over the batch. The matrix-based conditional entropy $S_{\alpha}(\mathbf{K}_{\bm{S}^{(i)}} | \mathbf{K}_{\bm{E}^{(i)}})$ quantifies the amount of uncertainty remaining for $\bm{S}^{(i)}$ (class label) after observing $\bm{E}^{(i)}$ (i.e. latent factors). Ideally, this quantity should be maximized so that observing exclusive information gleans nothing about shared information.\par

\begin{figure*}[!t]
    \centering
    \includegraphics[scale=0.4]{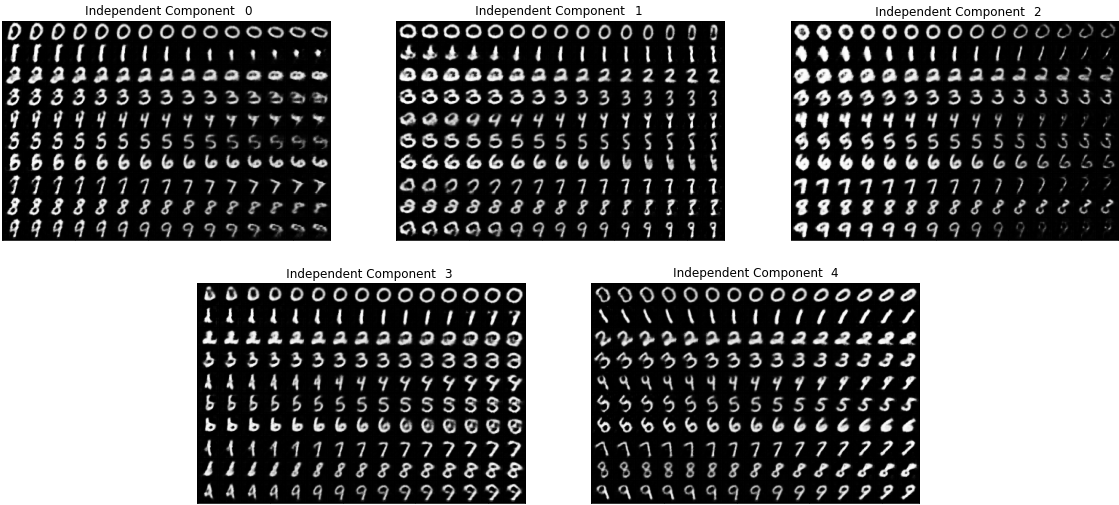}
    \caption{Walking on disentangled exclusive independent components. The walk is performed by encoding a digit prototype (center column) and modifying the exclusive dimensions. The leftmost and rightmost columns are moving -2 and +2 units, respectively, in the direction of the independent component, with evenly spaced steps in between. }
    \label{fig:ica_walk}
    \vspace{-6mm}
\end{figure*}

Here, we use the same encoder setup as used in Section~\ref{subsection:multiview MNIST}. To encourage the usage of exclusive dimensions by the encoder $f_i$, we also minimize reconstruction error by passing the full latent code through a decoder $g_{i}$. This is in contrast to Section \ref{subsection:multiview MNIST} where only an encoder is needed. The reconstruction of $\bm{X}^{(i)}$ is denoted as $\hat{\bm{X}}^{(i)} = g_i(f_i(\bm{X}^{(i)}))$. A separate encoder/decoder pair is used for each view. The full optimization problem is 
\begin{equation}
\begin{split}
\underset{f_1,g_1,f_2,g_2}{
\operatorname{maximize}} \quad & \textrm{DiME}_{\alpha}(\bm{S}^{(1)}, \bm{S}^{(2)}, \theta) + S_{\alpha}(\mathbf{K}_{\bm{S}^{(1)}}| \mathbf{K}_{\bm{E}^{(1)}}) + S_{\alpha}(\mathbf{K}_{\bm{S}^{(2)}}| \mathbf{K}_{\bm{E}^{(2)}}) \\& - \textrm{MSE}(\bm{X}^{(1)}, \hat{\bm{X}}^{(1)}) - \textrm{MSE}(\bm{X}^{(2)}, \hat{\bm{X}}^{(2)}).
\end{split}
\end{equation}
We use this to learn 10 shared dimensions and 5 exclusive dimensions. Unlike variational autoencoders, which also learn latent factors in an information-theoretic fashion, we are not enforcing that each exclusive dimension is independent of the others \citep{kingma2019introduction}, but instead letting them have dependence~\citep{cardoso1998multidimensional,hyvarinen2000emergence,casey2000separation,von2021self}. In order to visualize what exclusive factors are being learned, we calculate the independent components of the exclusive dimensions with  parallel FastICA~\citep{hyvarinen2000independent}. Using these independent components, we can start with a prototype and perform walks in the latent space in independent directions. By walking in a single independent direction, we intend to change only a single latent factor. In Figure \ref{fig:ica_walk}, we walk on the five exclusive independent components of View 1. Qualitatively, it seems that they encode height, width, stroke width, roundness, and rotation.

\subsection{Colored MNIST}
\begin{figure*}[!t]
     \begin{subfigure}[b]{0.07\linewidth}
         \centering
         \includegraphics[width=\textwidth]{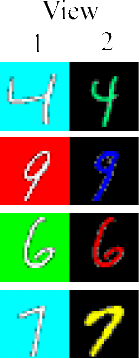}
         \caption{}
         \label{fig:colored dataset}
     \end{subfigure}
     \hfill
     \begin{subfigure}[b]{0.2\linewidth}
         \centering
         \includegraphics[width=\textwidth]{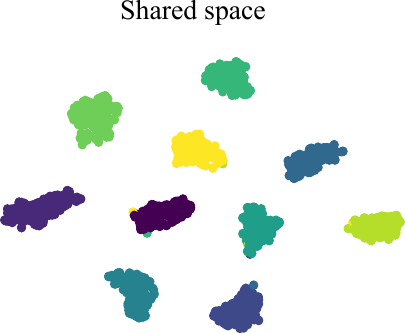}
         \caption{}
         \label{fig: shared space}
     \end{subfigure}
     \hfill
     \begin{subfigure}[b]{0.15\linewidth}
         \centering
         \includegraphics[width=\textwidth]{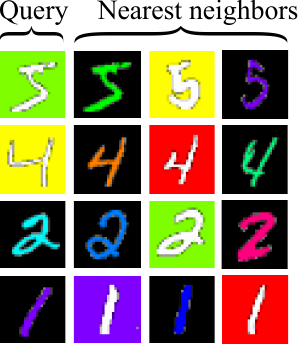}
         \caption{}
         \label{fig:nearest shared}
     \end{subfigure}
     \hfill
    \begin{subfigure}[b]{0.22\linewidth}
         \centering
         \includegraphics[width=\textwidth]{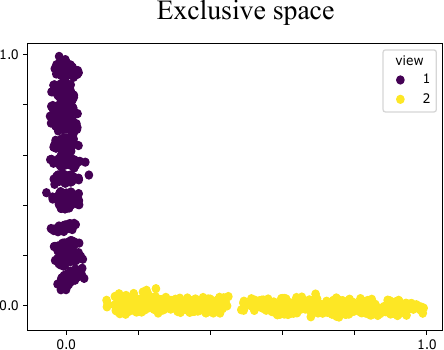}
         \caption{}
         \label{fig:exclusive space}
     \end{subfigure}
     \hfill
    \begin{subfigure}[b]{0.15\linewidth}
         \centering
         \includegraphics[width=\textwidth]{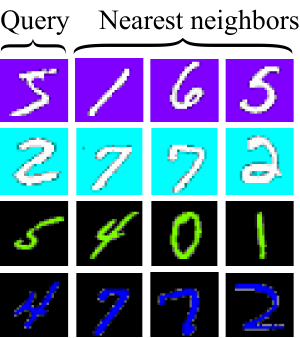}
         \caption{}
         \label{fig:nearest exclusive}
     \end{subfigure}
     \hfill
        \caption{(\textbf{a}) Example of a pair of images from the multiview colored MNIST: view 1 corresponds to background colored digits and view 2 to foreground colored digits. (\textbf{b}) t-SNE of the shared features learned via DiME. Colors represent different digit labels. (\textbf{c}) Image retrieval by finding nearest neighbors on the shared features. (\textbf{d}) view-exclusive features learned by minimizing JRD to a desired prior. (\textbf{e}) Image retrieval by finding nearest neighbors on the view-exclusive features}
        \label{fig:Results colored mnist}
\end{figure*}

Next we conduct experiments for learning disentangled representations by using DiME on the colored MNIST dataset, as in \citet{sanchez2020learning}. Given a pair of images from the same class but from different views (colored background and colored foreground digits, see Figure \ref{fig:colored dataset}), we seek to learn a set of shared features that represent the commonalities between the images and disentangle the exclusive features of each view. These representations are useful for certain downstream tasks, such as image retrieval by finding similar images to a query based on its shared or exclusive features. 
 
 We maximize the MI between the shared representations captured by a single encoder $f_{\textrm{sh}}$ and minimize the MI of the shared and exclusive features captured by a separate encoder $f_{\textrm{ex}}$ to encourage the disentanglement of the two components.
 In particular, $f_{\textrm{sh}}(\mathbf{X}) = \mathbf{S}$ encodes the shared features of the two views (class-dependent features), and $f_{\textrm{ex}}(\mathbf{X}) = \mathbf{E} = [\mathbf{V}, \mathbf{Z}]$ extracts the class-independent attributes so that $\mathbf{V}$ contains the exclusive features of each view (view-dependent features). $\mathbf{Z}$ is used to encode residual latent factors needed for reconstruction.\par

Our training procedure consists of two steps to follow the work of \citet{sanchez2020learning}. First, because we want $f_{\textrm{sh}}$  to learn a shared space that is invariant to the view, we first maximize the MI between the shared representations of both views $ \mathbf{S}^{(1)}$ and $ \mathbf{S}^{(2)}$ via DiME as in \eqref{eq:dime_multiview}. The same encoder $f_{\textrm{sh}}$ is used for both views. Second, once the shared representation is learned, we freeze the shared encoder and train the exclusive encoder $f_{\textrm{ex}}$. To disentangle the three latent subspaces, we minimize the matrix-based MI between $\mathbf{S}$ and  $\mathbf{E}$ and minimize the matrix-based MI between $\mathbf{V}$ and $\mathbf{Z}$.\footnote{We do this minimization with matrix-based MI, rather than DiME, because DiME is a lower-bound on matrix-based MI and may not be suited for minimization tasks.} This procedure encourages the exclusive encoder not to learn features that were already learned by the shared encoder. Additionally, we place a conditional prior on $\mathbf{V} = [\mathbf{V}_1,\mathbf{V}_2] $ such that  $\mathbf{V}^{(1)} = [\mathbf{V}_1, \mathbf{0}] $ and $\mathbf{V}^{(2)} = [\mathbf{0},\mathbf{V}_2] $ intending different subspaces to independently encode the view-exclusive generative factors (see Figure \ref{fig:exclusive space}). We achieve this by minimizing the Jensen-Rényi divergence (JRD) \citep{osorio2022representation} of the view-exclusive features to samples from two mutually exclusive uniform random variables from 0 to 1, which we denote as $\mathbf{P}$. To ensure that features have high MI with the images, we also train a decoder $g$ that will reconstruct the original images using all three latent subspaces. We train the exclusive encoder by optimizing  $f_{\textrm{ex}}, g $ to minimize the loss function $    \mathcal{L}_{\textrm{ex}} = I_\alpha(\mathbf{S},\mathbf{E}) + I_\alpha(\mathbf{V},\mathbf{Z})  + D_{\alpha} (\mathbf{V}||\mathbf{P}) + \textrm{MSE}(\mathbf{X},\hat{\mathbf{X}})$. Exact model details are provided in Appendix \ref{appendix:colored-mnist-model}.

A t-SNE visualization of the learned shared features (Figure \ref{fig: shared space})  shows how digits from the same class are grouped into the same clusters regardless of the view. This is corroborated by image retrieval by finding the nearest neighbors to the shared representation of a query image (Figure~\ref{fig:nearest shared}). DiME captures the common attributes between the two views, which in this case is class information. On the other hand, the learned view-exclusive features (Figure \ref{fig:exclusive space}) are class-agnostic and for a given query, the exclusive space nearest neighbors correspond to the same background/foreground color. This is independent of the digit class (Figure \ref{fig:nearest exclusive}). To further support the quality of disentanglement, we give a visualization in Appendix \ref{appendix:color-mnist-styletransfer} where we perform style transfer.\par

\begin{table}[!]
\begin{subtable}{.5\textwidth}
\centering
\scalebox{0.75}{
\begin{tabular}{l|l|l|l}
Method                             & DN     & B & F \\ \hline
Ideal                              & 100\%            & 8.33\%           & 8.33\%           \\
S. 2020   & 94.48\%          & \textbf{8.22\%}  & \textbf{8.83}\%           \\
GG. 2018 & 95.42\%          & 99.56\%          & 29.81\%          \\
MINE & \textbf{98.92}\% & 11.24 \% &  12.11\%  \\
DiME                               & \textbf{98.93\%} & \textbf{8.79\%  }         & \textbf{8.41\%} 
\end{tabular}
}
\end{subtable}
\begin{subtable}{.5\textwidth}
\centering
\scalebox{0.75}{
\begin{tabular}{l|l|l|l}
Method                             & DN     & B & F \\ \hline
Ideal                              & 10.00\%            & 100\%           & 100\%           \\
S. 2020   & 13.20\%          & \textbf{99.99\%}  & \textbf{99.92}\%           \\
GG. 2018 & 99.99\%          & 71.63\%          & 29.81\%          \\
MINE   & \textbf{10.85\%}          & 69.38\%  & 70.24\%           \\
DiME              & \textbf{10.30\%} & 98.84\%           & 84.3\%
\end{tabular}
}
\end{subtable}%
\caption{Digit number (DN), Background (B) and Foreground (F) color accuracy using the (\textbf{left}) shared representations and  (\textbf{right}) exclusive representations. Note that S. 2020 refers to \citet{sanchez2020learning} and GG. 2018 refers to \citet{gonzalez2018image}.}
\label{tab:accuracy}
 \vspace{-5mm}
\end{table}
As in \citet{sanchez2020learning}, we evaluate the disentanglement of the shared and exclusive features by performing some classification experiments. Ideally, a digit classifier trained on the shared features should accurately classify the digit label with poor color classification. Conversely, the classification of exclusive features should perform well in identifying color and randomly for digit labels. We train a classifier and compare our results to \citet{sanchez2020learning}, \citet{gonzalez2018image}, and MINE in Table \ref{tab:accuracy}. DiME achieves the highest digit number classification accuracy on shared features. Additionally, the foreground and background color accuracy is close to a random guess, verifying the disentanglement between shared and exclusive features. For the MINE baseline, we replaced DiME with MINE while keeping the rest of the losses. When using MINE (or other MI estimators), it is necessary to tune the kernel bandwidth for matrix-based quantities. On the contrary, DiME naturally adjusts the data with a fixed bandwidth ($\sqrt{D/2}$ was used) to make the representations suitable for matrix-based quantities. For MINE, we tested 5 different values of kernel bandwidth selected in a $\log$ space around $\sqrt{D/2}$ and show the best results.
\section{Conclusions}
\label{section:conclusion}
We proposed DiME, a quantity that behaves like MI and can be estimated directly from data. DiME is built using matrix-based entropy, which uses kernels to measure uncertainty in the dataset. We compared the behavior of DiME as an MI estimator to variational estimators and showed it was well-behaved for sufficiently large batch sizes. Unlike variational estimators, DiME does not require a critic network and is calculated in the same space as data representations. We applied DiME to the problems of multiview representation learning and factor disentanglement. To handle the variance of DiME, which can be very large, we are exploring alternatives to the Gram matrix formulation. For instance, by working directly in the feature space, we can aggregate results from small batches and reduce the variance of the estimators. There are several methods to provide an explicit feature space, such as using Random Fourier Feature approximations~\citep{rahimi2007random} to kernels.

\section{Limitations}
Below are some limitations of this work:

\begin{itemize}
    \item \textbf{Sensitivity to Batch Size} As pointed out in Section \ref{sect:variational-bounds}, one method of reducing the variance of DiME is by increasing the batch size. However, we acknowledge that not all datasets are amenable to large batch sizes. In such cases, other methods can be used to reduce the variance of DiME such as increasing the number of permutations used to approximate the expectation.
    \item \textbf{Minimization of Mutual Information} Since the self-regulation of DiME comes from maximizing the difference of entropies, applications that seek to minimize mutual information would require solving a $\operatorname{minmax}$ optimization. In such cases, alternatives to DiME seem preferable.
    \item \textbf{Estimation of Shannon's Mutual Information} As mentioned previously, DiME is not an estimator of Shannon's MI. It shares several properties with Shannon's MI that allow it to be used, for instance, in tasks of maximizing MI. However, DiME is not able to be used for tasks that require actual estimation of Shannon's MI where measurement units (bits or nats) are important.
\end{itemize}

\section{Acknowledgements}
This material is based upon work supported by the Office of the Under Secretary of Defense for Research and Engineering under award number FA9550-21-1-0227.

\bibliography{main}

\newpage
\appendix
\section{Behavior of DiME with respect to Kernel Bandwidth}
\label{appendix:dime-bandwidth-behavior}
\begin{figure}[!h]
    \centering
    \subfloat{{\includegraphics[scale=0.6]{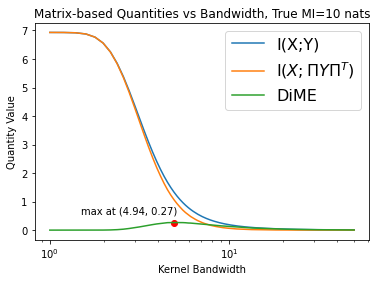}}}%
    
\subfloat{{\includegraphics[scale=0.6]{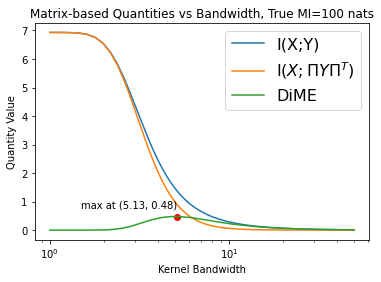}}}%
    \caption{Comparison of DiME and matrix-based MBMI as kernel bandwidth $\sigma$ is changed. $X$ and $Y$ are drawn from the correlated Gaussian dataset mentioned in Section \ref{sect:variational-bounds}. $X$ and $Y$ have $n=1024$ samples each. The Shannon MI between $X$ and $Y$ is \textbf{(top)} 10 nats and \textbf{(bottom)} 100 nats.}
    \label{fig:bandwidth_behavior}
\end{figure}

In Section \ref{sect:dime-intro}, we argue that DiME "moves from small to large and back to small as we decrease the kernel bandwidth $\sigma$ from $\sigma=\infty \rightarrow \sigma=0$". In other words, there exists a maximal value of DiME for some finite $\sigma \neq 0$. This is in contrast to matrix-based mutual information which achieves a maximum value of $\ln(n)$ when $\sigma=0$ due to saturation. These behaviors are shown in Figure $\ref{fig:bandwidth_behavior}$.

\newpage
\section{Further Details of Variational Bounds Comparisons}

\subsection{Alternative Plot for Figure \ref*{fig:variationalbounds}}
\label{appendix:alternative-value-plot}
\begin{figure*}[!h]
     \centering
     \begin{subfigure}[b]{\textwidth}
         \centering
         \includegraphics[width=\textwidth]{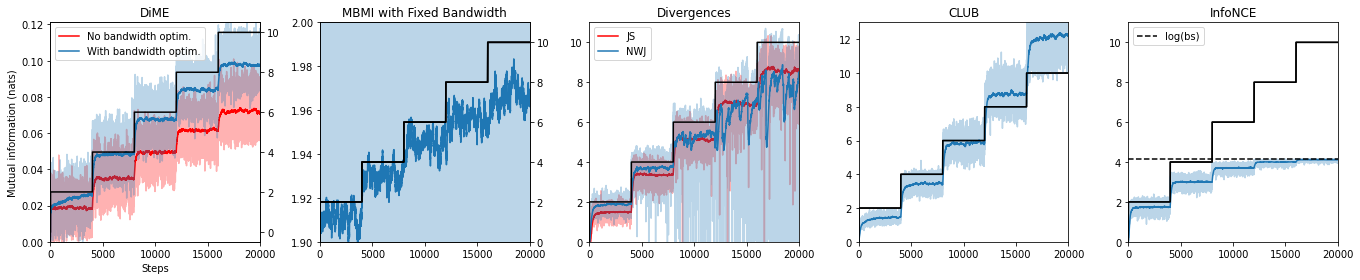}
         \label{fig:value_variationalbounds_64}
     \end{subfigure}
     \centering
     \begin{subfigure}[b]{\textwidth}
         \centering
         \includegraphics[trim=0 40 0 40, width=\textwidth]{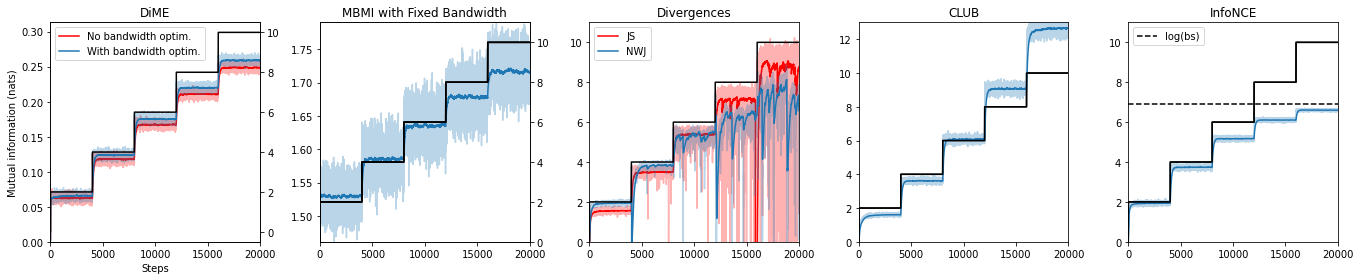}
         \label{fig:value_variationalbounds_256}
     \end{subfigure}
     \caption{Performance of different estimators of MI for a toy correlated Gaussian dataset with increasing correlation over time. The black staircase line denotes the true MI of the dataset. \textbf{(top)} Trained with batch size of 64 and using joint critics for variational estimators. \textbf{(bottom)} Trained with batch size of 1024 and using separable critics for variational estimators. }
    \label{fig:value_variationalbounds}
\end{figure*}

We showed in Figure \ref{fig:variationalbounds} a relative comparison of DiME to variational MI estimators. This relative comparison shows how changes in the underlying true MI proportionally affect the estimators. We chose to show relative quantities, rather than the actual quantity values, because matrix-based information-theoretic quantities are on a different scale than the true MI and the other estimators. Here in Figure \ref{fig:value_variationalbounds}, we show an alternative version of the plot which depicts the actual quantity values. Note that there are two different y-axes in some graphs: the left y-axis shows the quantity values and the right y-axis shows the true MI.

\subsection{Behavior of DiME with respect to Batch Size and Dimensionality}
\label{appendix:dime-batch-behavior}

\begin{figure}[!ht]
     \centering
     \begin{subfigure}{0.8\textwidth}
         \centering
         \includegraphics[width=\textwidth]{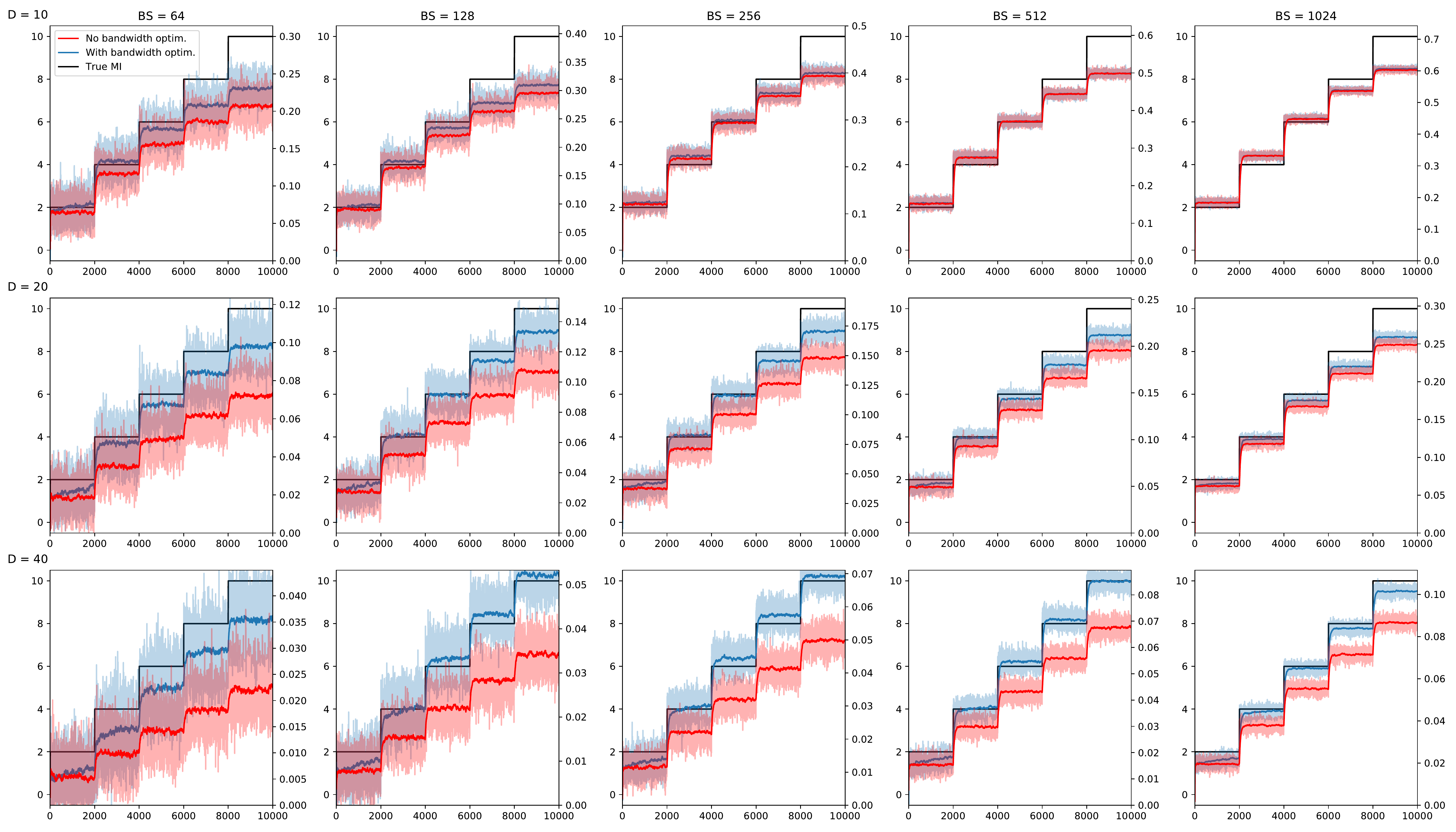}

         \caption{For varying batch sizes (columns) and dimensionalities (rows), we use DiME to approximate the MI of a toy dataset. We use a Gaussian kernel as DiME's kernel function. Learned kernel bandwidths are plotted below. For the fixed kernel bandwidths, we use $\sigma = \sqrt{D/2}$}
        \label{fig:mi_biggrid}
        
     \end{subfigure}
     
     \begin{subfigure}{0.8\textwidth}
         \centering
         \includegraphics[width=\textwidth]{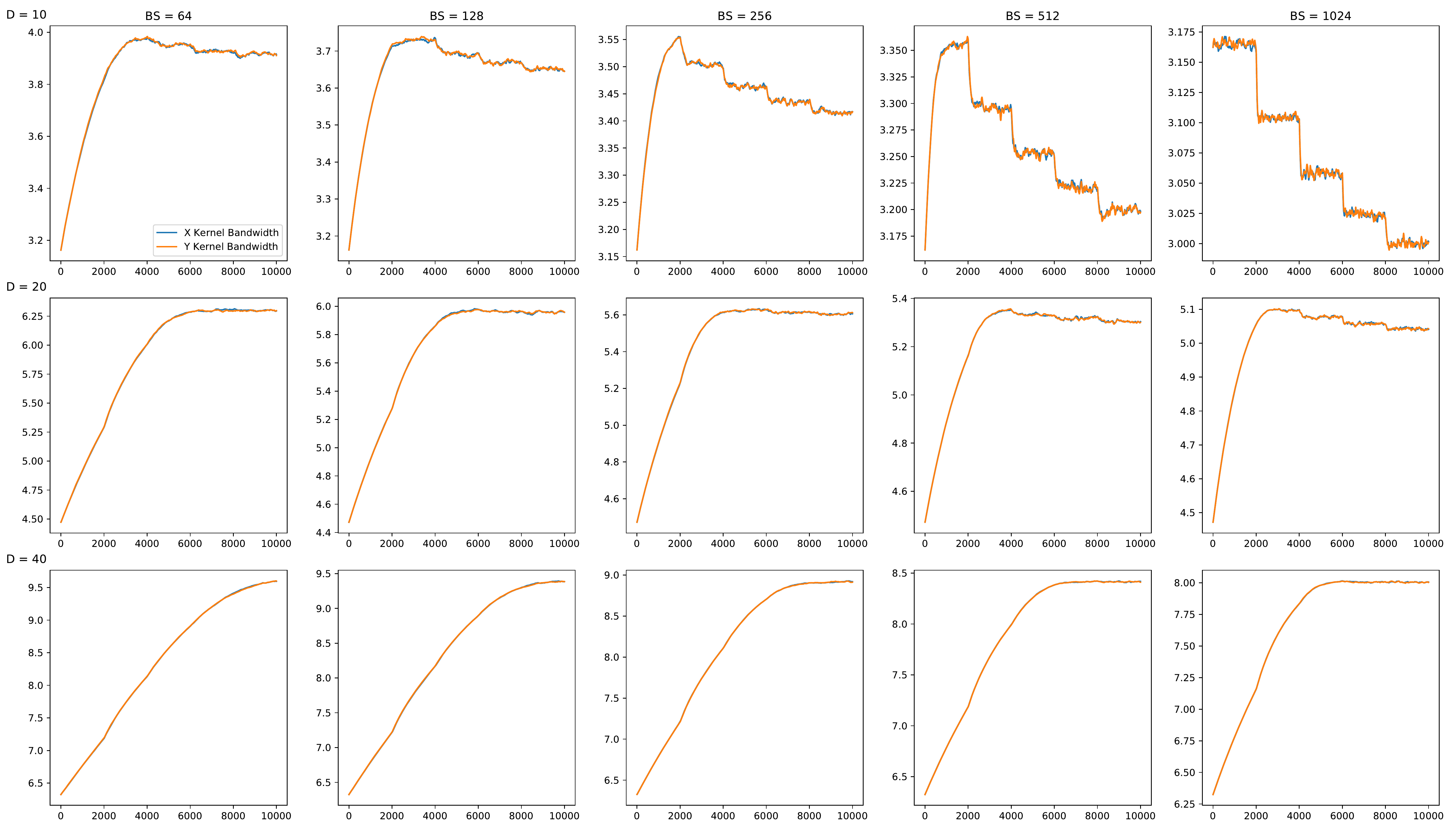}
         \caption{Values of learned kernel bandwidths. Each plot corresponds to the plot in the same position in the above figure.}
         \label{fig:sigma_biggrid}
         
     \end{subfigure}
     \caption{Behavior of DiME as data dimensionality and batch size are varied.}
    \label{fig:dime_biggrid}
\end{figure}

We display how DiME behaves by changing the batch size and dimensionality in Figures \ref{fig:mi_biggrid} and \ref{fig:sigma_biggrid}. We use the same toy dataset (from \cite{poole2019variational}) as described in Section \ref{sect:variational-bounds}. Note that the true MI is on a different scale from DiME, and the DiME y-axis is on the right-hand side of each graph. Here we use only one random permutation to approximate the expectation in DiME. The darker lines are weighed running averages with a window size of 100. A few critical behaviors can be observed in \ref{fig:mi_biggrid}:

\begin{enumerate}
    \item Increasing batch size decreases variance and increases the bias
    \item Increasing dimensionality increases variance and decreases the bias
    \item Kernel bandwidth optimization has a larger effect in high dimensionalities
\end{enumerate}

These behaviors are important to keep in mind when using DiME. It may be possible to normalize DiME in such a way that helps stabilize its bias with regard to changes in batch size or dimensionality, but we leave that to future work.

\newpage

\section{Further Details of Multiview MNIST Experiments}

\subsection{Multiview MNIST Model Details}
\label{appendix:multiview-model}
Here we describe the model architecture used in Sections \ref{subsection:multiview MNIST} and \ref{subsection:disentanglement}. 

In Section \ref{subsection:multiview MNIST}, we trained the encoder architecture described in Table \ref{tab:multiview_architecture}. To test downstream accuracy on the rotated view,
we used a linear classifier with one hidden layer mapping (D → 1024 → 10). Separate encoders and classifiers were used for each view. These architectures were used for DiME and all baselines. We used Adam to train with a learning rate of
0.0005, batch size of 3000, for 100 epochs. The latent dimensionality D was ranged between 1 and 15. For the DiME and
CKA objective functions, we used the Gaussian kernel  with a fixed kernel parameter of $\sqrt{D/2}$.

In Section \ref{subsection:disentanglement}
, we used both the encoder and decoder described in Table \ref{tab:multiview_architecture}. We used Adam to train with a learning rate of
0.0005, batch size of 3000, for 100 epochs. We used DiME as described in the preceding paragraph. We use a shared latent
dimensionality of 10 and an exclusive dimensionality of 5. Thus the total latent dimensionality (D in Table \ref{tab:multiview_architecture}) is 15. Note
that DiME only operates on the 10 shared latent dimensions from each view, while conditional entropy operates on both
shared latent dimensions and exclusive dimensions. Additionally, the decoder uses the entire 15 latent dimensions.

\begin{table}[t]
\centering

    \caption{Description of the architectures employed in multiview MNIST experiments}
    
    \label{tab:multiview_architecture}
\scalebox{0.85}{
    \begin{tabular}{l|l}
    \textbf{Encoder} & \textbf{Decoder} 
    \\\hline
    Input: $28\times 28 \times 1$ & Input: $1\times 1 \times D$ \\ 
    
    $3\times3$ conv, $8$ out channel, stride $2$, padding $1$ & Linear $D$ in dimensions, $1024$ out dimensions\\
    
    ReLU & ReLU \\
    
    $3\times3$ conv, $16$ out channel, stride $2$, padding $1$ & Linear $1024$ in dimensions, $3*3*32$ out dimensions \\
    
    Batch Normalization & reshape to $32 \times 3 \times 3$ \\
    ReLU & $3\times3$ convTrans, $16$ out channel, stride $2$ \\
    
    $3\times3$ conv, $32$ out channel, stride $2$ & 
    Batch Normalization \\
    
    ReLU & ReLU \\
    
    LazyLinear, $1024$ out dimensions & $3\times3$ convTrans, $8$ out channel, stride $2$, padding=$1$, output padding=$1$\\
    Linear, $1024$ in dimensions, $D$ out dimensions & Batch Normalization  \\
    & ReLU\\
    & $3\times3$ convTrans, $1$ out channel, stride $2$, padding=$1$, output padding=$1$
    \end{tabular}
    }
\end{table}

\section{Further Details of Colored MNIST Experiments}
\label{appendix:colored-mnist-model}
\subsection{Colored MNIST Model Details}
For the shared encoder $f_{sh}$ we train the architecture shown in Table \ref{tab:coloredmultiview_architecture}. Here, we used Adam with a learning rate of 0.0001 and a batch size of 1500 for 50 epochs to learn 10 features. Once this model is trained, we freeze it and learn the exclusive encoder $f_{ex}$ and the decoder $g$  with Adam with a learning rate of 0.0001, batch size of 64 for 100 epochs. The architecture of $f_{ex}$ is the same used for $f_{sh}$ and the decoder architecture is also shown in \ref{tab:coloredmultiview_architecture}. The dimensionality of $ \mathbf{S}, \mathbf{V}$ and $\mathbf{Z}$ are 10, 2 and 6 respectively.

\begin{table}[!h]
    
        \caption{Description of the architectures employed in the colored multiview MNIST experiments}
    \label{tab:coloredmultiview_architecture}
    \resizebox{\linewidth}{!}{
    \begin{tabular}{l|l}
    \textbf{Encoder} & \textbf{Decoder} 
    \\\hline
    Input: $28\times 28 \times 3$ & Input: $1\times 1 \times D$ \\ 
    
    $3\times3$ conv, $32$ out channel, stride $1$, padding $1$ & Linear $D$ in dimensions, $1024$ out dimensions\\
    
    ReLU & ReLU \\
    
    $3\times3$ conv, $64$ out channel, stride $1$, padding $1$ & Linear $1024$ in dimensions, $4\times4\times256$ out dimensions \\
    ReLU & ReLU\\
    2D Maxpooling  $2\times 2$ & $3\times3$ convTrans, $256$ out channel, stride $2$, padding=$1$  \\
    
    $3\times3$ conv, $128$ out channel, stride $1$, padding $1$ & ReLU\\
    ReLU & $3\times3$ conv, $128$ out channel, stride $1$, padding $1$\\
    $3\times3$ conv, $128$ out channel, stride $1$, padding $1$ & ReLU \\
    ReLU & $3\times3$ conv, $128$ out channel, stride $1$, padding $1$\\
    2D Maxpooling  $2\times 2$ & ReLU  \\
    $3\times3$ conv, $256$ out channel, stride $1$, padding $1$ & $4\times4$ conv, $64$ out channel, stride $2$, padding $1$\\
    ReLU & ReLU\\
    2D Maxpooling  $2\times 2$ &  $3\times3$ conv, $32$ out channel, stride $1$, padding $1$\\
    Linear, $144\times4\times4$ in dimensions, 1024 out dimensions & ReLU\\
    Linear, 1024 in dimensions, D out dimensions & $4\times4$ conv, $3$ out channel, stride $2$, padding $1$
    \end{tabular}}

\end{table}

\newpage

\subsection{Disentanglement of the residual latent factors}
\label{appendix:color-mnist-disentangelment}
\begin{figure}[!b]
     \centering
     \begin{subfigure}[b]{0.8\textwidth}
         \centering
         \includegraphics[width=\textwidth]{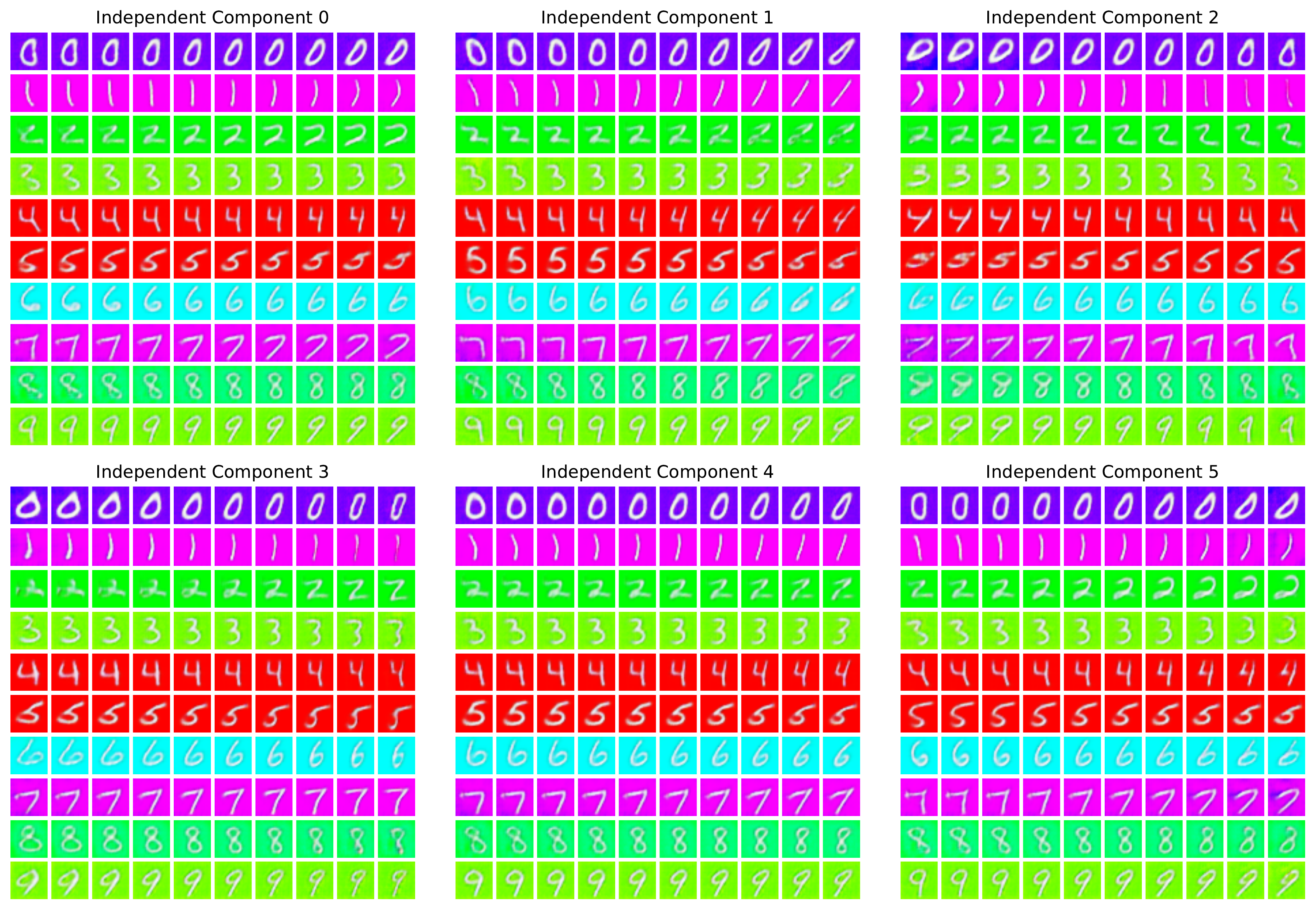}
         \label{fig:ica_walk_v1}
        \caption{ICA walks on View 1}
     \end{subfigure}
     \hfill
     \centering
     \begin{subfigure}[b]{0.8\textwidth}
         \centering
         \includegraphics[width=\textwidth]{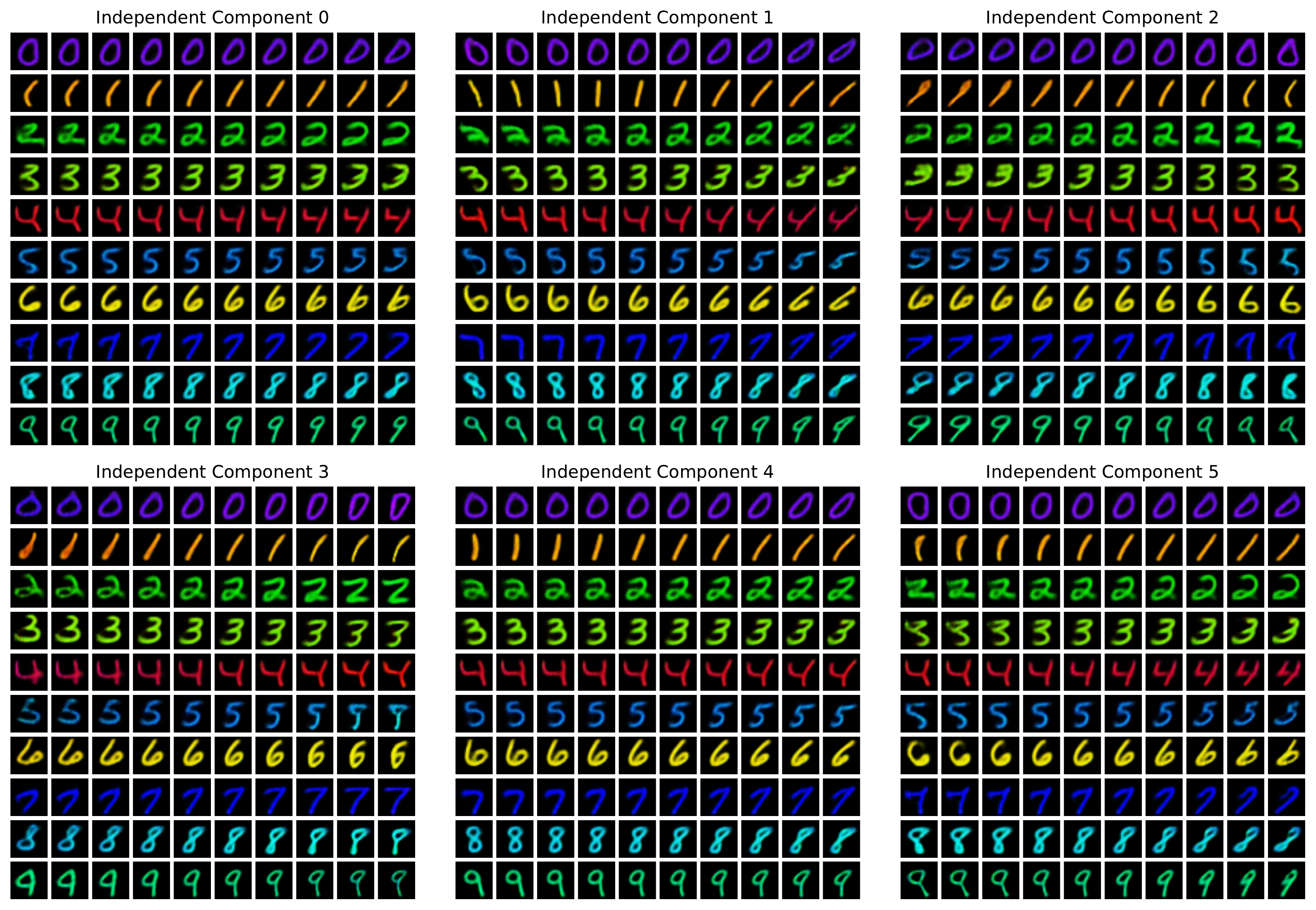}
         \label{fig:ica_walk_v2}
        \caption{ICA walks on View 2}
     \end{subfigure}
     
    \caption{Walking on the independent components of the residual latent factors of the colored MNIST dataset.}
    \label{fig:icaColored}
\end{figure}

To evaluate what the residual factors $\mathbf{Z}$ are capturing, we extracted their independent components to see if they capture different aspects of the digits. In Figure \ref{fig:icaColored} we walk on the six independent components which seem to encode different prototypes of the digits, except for the independent component 1 which clearly encodes the rotation of the digit.  

\newpage
\subsection{Style Transfer}
\label{appendix:color-mnist-styletransfer}
\begin{figure}[!h]
     \begin{subfigure}[b]{0.3\linewidth}
         \centering
         \includegraphics[width=\textwidth]{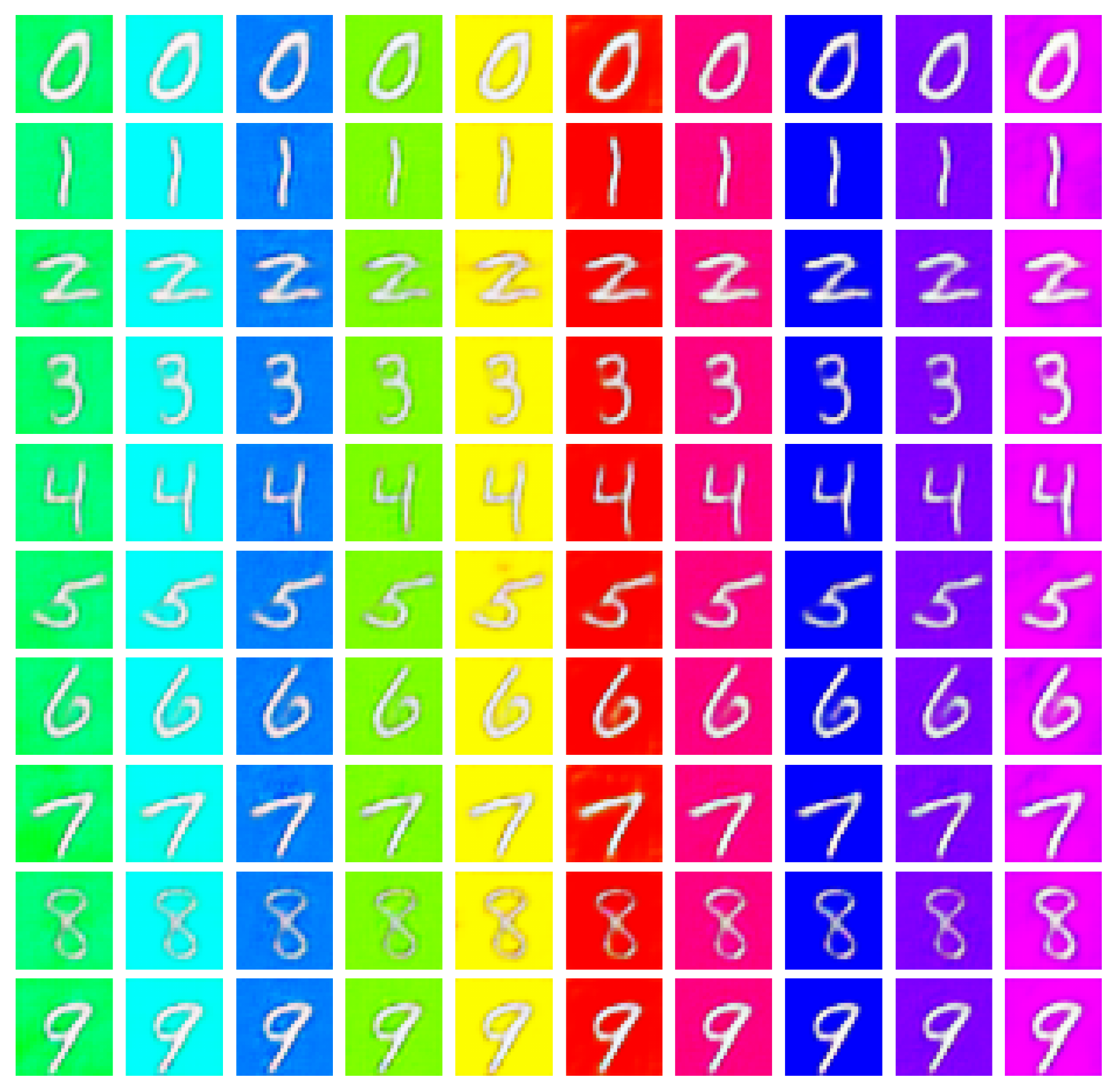}
         \caption{}
         \label{fig:Background disentangled}
     \end{subfigure}
     \hfill
     \begin{subfigure}[b]{0.3\linewidth}
         \centering
         \includegraphics[width=\textwidth]{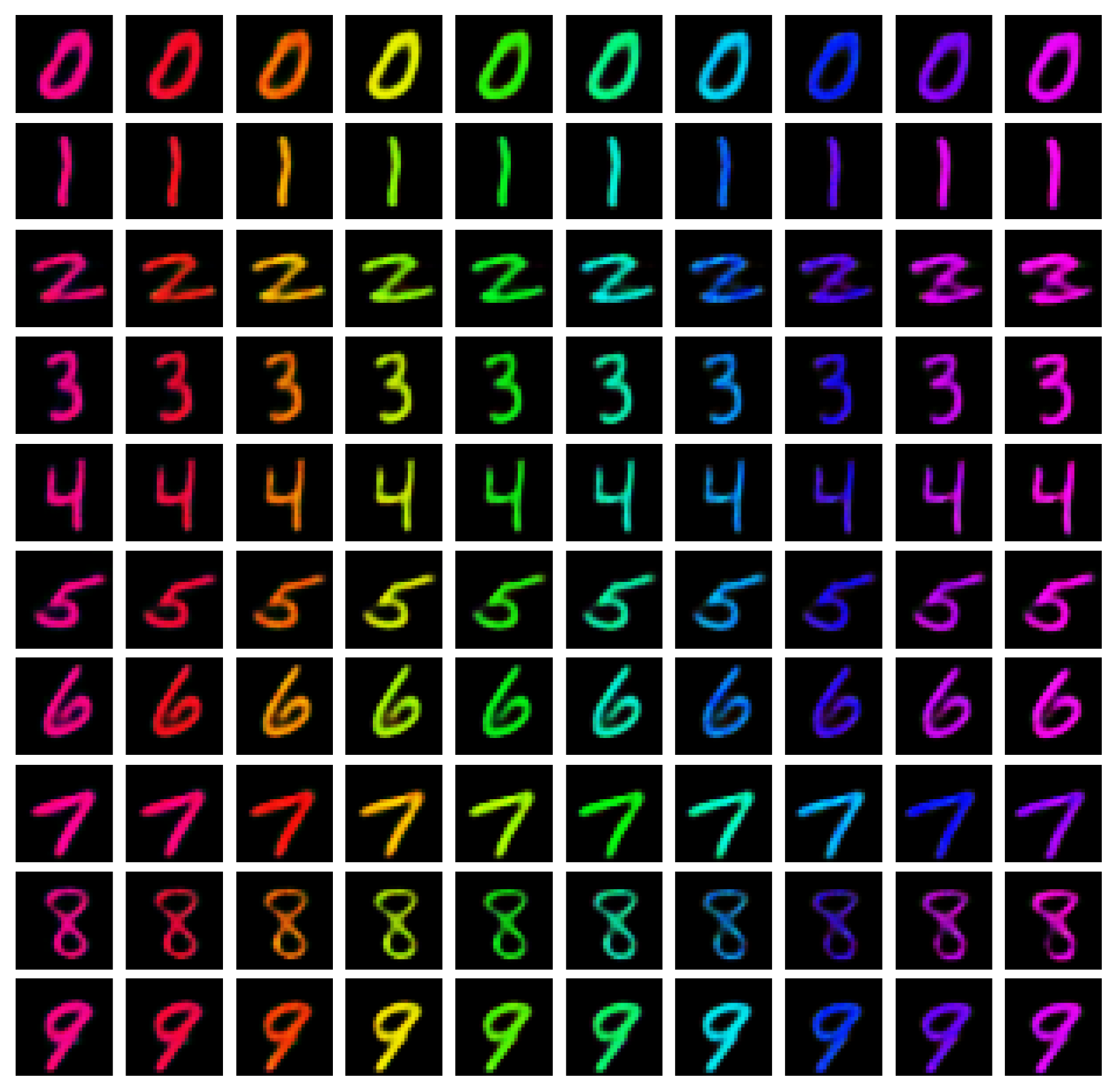}
         \caption{}
         \label{fig:foreground_disentangled}
     \end{subfigure}
     \hfill
     \begin{subfigure}[b]{0.09\linewidth}
         \centering
         \includegraphics[width=\textwidth]{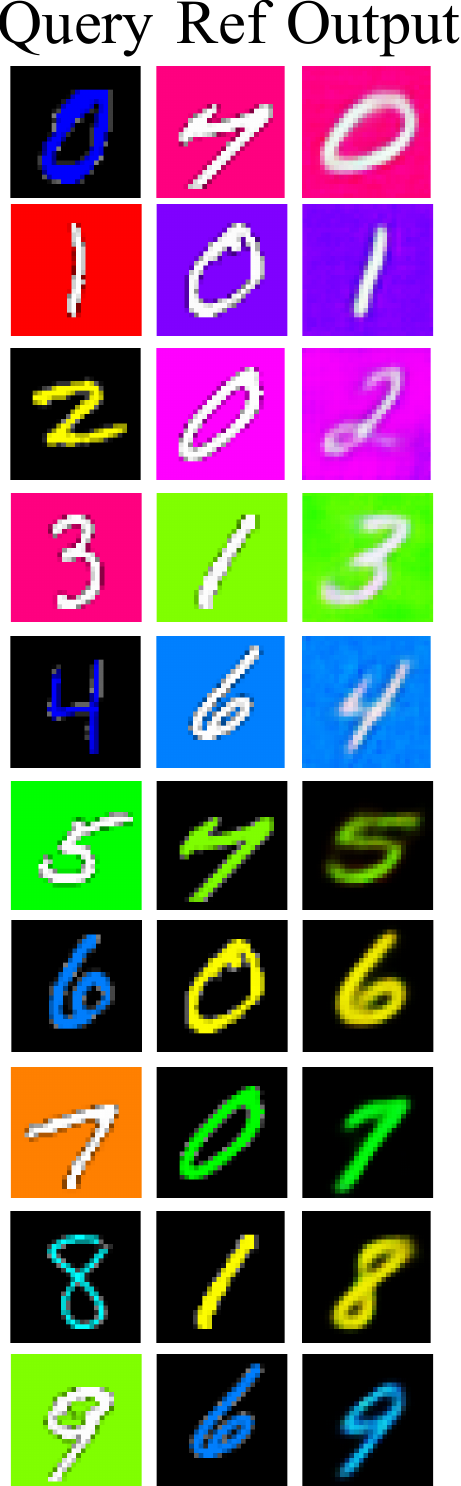}
         \caption{}
         \label{fig:style_transfer}
     \end{subfigure}
     \hfill
        \caption{Disentangling of generative factors in the multiview colored MNIST dataset. (a) Walking on the exclusive dimension of view 1. (b) Walking on the exclusive dimension of view 2. (c) Style transfer results by swapping exclusive dimensions.}
        \label{fig:colored mnist results}

\end{figure}

To further support the quality of the shared features learned by DiME, we generate samples by keeping the shared representation fixed and moving along the view-exclusive dimensions. In Figures \ref{fig:Background disentangled} and \ref{fig:foreground_disentangled} we can see how the digit content is well preserved as the view generative factor is changing. Specifically, each row corresponds to a random digit image selected from the dataset. We encode this image, modify the view-exclusive dimension, and visualize the decoded image.

This approach is also well suited for style transfer, by exchanging the style (exclusive $E$) between a query and a reference image while keeping the content (shared $S$) intact. Without any additional training required, we display style transfer in Figure \ref{fig:style_transfer}.

\section{DiME-GANs}
Motivated by the relation between MI  and Jensen-Shannon divergence, we propose training of GANs as an alternating optimization between two competing objectives based on DiME. The purpose of this is not to break new ground in GANs, but to instead highlight the versatility of DiME. Similar to MMD-GAN \cite{Li2017mmdgan}, our discriminator network $f_\theta$ maps samples from data space $\mathcal{X}$ to a representation space $\mathcal{Y}$, and the generator network $g_\psi$ maps samples from the noise space $\mathcal{Z}$ to data space $\mathcal{X}$. \par
To train the discriminator, we use DiME to maximize the MI between samples of a mixture distribution and an indicator variable. The mixture distribution has two components: the distribution of real samples and the distribution of samples from a generator. The indicator variable is paired with each sample drawn from the mixture and indicates the component (real or fake) from which the sample came from. To train the generator, we maximize the matrix-based conditional entropy of the indicator variable given a sample from the mixture. This is equivalent to minimizing the matrix-based MI between the mixture and the indicator variable. Since the matrix-based MI is an upper bound of DiME, each generator update tries to reduce DiME by pushing down this upper bound. Conversely, each discriminator step tunes the representation space such that the upper bound given by the matrix-based MI is tight.

\begin{figure}[!b]
    \centering
    \includegraphics[width=0.75\linewidth]{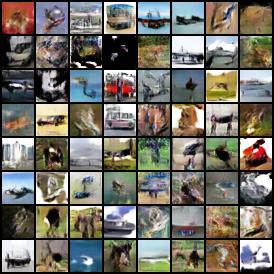}
    \caption{Samples generated using DiME-GAN}
    \label{fig:GAN_cifar10}
    \vspace*{-5pt}
\end{figure}

Figure \ref{fig:GAN_cifar10} shows samples from DiME-GAN trained on CIFAR10 dataset. For the kernel we use the Laplacian kernel defined as $\kappa_{\sigma}(x, y) = e^{-\frac{\sum_{i=1}^D \vert (x)_i - (y)_i\vert}{\sigma}}$ with sigma fixed $\sigma = \sqrt{D/2}$. The discriminator objective tunes the parameters of $f_\theta$. For each iteration, we use 64 images from the true distribution and 64 images sampled from the generator network. Both optimizers use Adam with $lr = 0.00005$. The architecture is similar to DCGAN, but without batch-normalization and gradient clipping during training. Details of the employed architecture and hyperparameters are provided in table \ref{tab:dimegan_architecture}.   

\newpage
\subsection{Details of the GAN and Objective Function}

We have two mappings: a discriminator $f_{\theta}$ network that maps points from data space $\mathcal{X}$ to representation space $\mathcal{Y}$, and a generator network $g_{\psi} $that maps noise samples from $\mathcal{Z}$ to data space $\mathcal{X}$. Namely, for CIFAR10, $\mathcal{X} \subset \mathbb{R}^{32\cdot 32\cdot 3}$, $\mathcal{Z} \subseteq \mathbb{R}^{D_Z}$, and $\mathcal{Y} \subseteq \mathbb{R}^{D_Y}$. Details of the architecture of the discriminator and generator networks are provided in Table \ref{tab:dimegan_architecture}

\begin{figure}[!h]
     \centering
     \begin{subfigure}{0.3\textwidth}
         \centering
         \includegraphics[width=\textwidth]{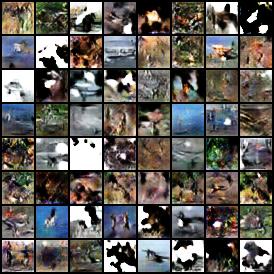}
         \label{fig:dimegan_gaussian}
         \caption{Gaussian kernel}
     \end{subfigure}
     \begin{subfigure}{0.3\textwidth}
             \centering
         \includegraphics[width=\textwidth]{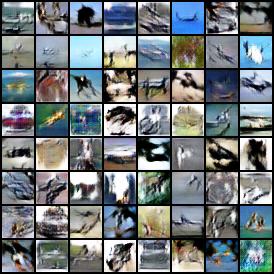}
         \label{fig:dimegan_elliptical}
         \caption{Elliptical Laplacian kernel}
     \end{subfigure}
     \begin{subfigure}{0.3\textwidth}
             \centering
         \includegraphics[width=\textwidth]{imgs/samples_dcgan_laplacian64_101022.jpeg}
         \label{fig:dimegan_laplacian}
         \caption{Factorized Laplacian kernel}
     \end{subfigure}
     \caption{Generated Images for DiME-GAN trained with different kernels}
    \label{fig:dimegan_generated_imag}
\end{figure}

Each full iteration of the algorithm is comprised of two updates: an update for the discriminator, and an update for the generator. Let $\mathbf{X}_{r}$ be the size $N_{\textrm{batch}}$ batch of real samples, $\mathbf{X}_f = g_{\psi}\left(\mathbf{Z}\right)$ a batch of generated samples, and $\mathbf{X} = \operatorname{cat}\{\mathbf{X}_r, \mathbf{X}_f\}$ a concatenated samples. Associated to $\mathbf{X}$ we have the label vector $\mathbf{l}$ that indicates which samples in $\mathbf{X}$ come from $\mathbf{X}_r$ and $\mathbf{X}_f$, respectively.     
The discriminator update is based on DiME, which measures the dependence between $\mathbf{X}$ and $\mathbf{l}$. At each iteration, we update $\theta$ to increase the difference of entropies.
\begin{equation}\label{eq:dime_gan_disc_update}
 \E_{\bm{\Pi}}\left[ S_{\alpha}(\mathbf{K}_{\mathbf{X}}(\theta) \circ \bm{\Pi}\mathbf{K}_{\mathbf{l}} \bm{\Pi}^{T})\right]\\ - S_{\alpha}(\mathbf{K}_{\mathbf{X}}(\theta) \circ \mathbf{K}_{\mathbf{l}}),
\end{equation}
where $\left( \mathbf{K}_{\mathbf{l}} \right)_{ij} = 1$ if $i$ and $j$ are both real or fake images, and $0$ otherwise. The kernel between two images corresponds to the composition of the discriminator mapping $f_{\theta}$ followed by a positive definite kernel $\kappa: \mathcal{Z}\times \mathcal{Z} \mapsto \mathbb{R}$, namely, $\left(\mathbf{K}_{\mathbf{X}}(\theta)\right)_{ij} = \kappa(f_{\theta}(x_i), f_{\theta}(x_j))$. 
The generator update is based on matrix-based conditional entropy of $\mathbf{l}$ given $\mathbf{Y} = f_{\theta}(\mathbf{X})$. In this step, the parameters of the discriminator remain fixed and the parameters $\psi$ of the generator are updated so that the conditional entropy,
\begin{equation}\label{eq:dime_gan_gen_update}
S_{\alpha}(\mathbf{K}_{\mathbf{Y}}(\psi) \circ \mathbf{K}_{\mathbf{l}}) - S_{\alpha}(\mathbf{K}_{\mathbf{Y}}(\psi)),
\end{equation}
increases. In this case, the gradients of the objective backpropagate to the generator network, since $\mathbf{Y}_{f} = f_{\theta}(g_{\psi}(\mathbf{Z}))$. To compute the Gram matrix, we use the same kernel $\kappa$ as in \eqref{eq:dime_gan_disc_update}. To optimize parameters, we use Adam with learning rate $lr = 0.00005$ and $\beta_1 = 0.5$. Note that there is one optimizer for the discriminator and another for the generator. For the kernel $\kappa$ we tried several options, listed in Table \ref{tab:kernels}.  In all experiments, we set the kernel parameter to $\sigma = \sqrt{D/2}$, for $x\in \mathbb{R}^{D}$. Figure \ref{fig:dimegan_generated_imag} shows examples of generated images after training the GAN with the DiME objective for the kernels described in Table \ref{tab:kernels}.

\begin{table}[!h]
    \centering
        \caption{Kernel functions that we tried in DiME-GAN experiments}
    \label{tab:kernels}
    \begin{tabular}{c|c}
    \textbf{Kernel}  & $\kappa(x_i,x_j)$ \\
    Gaussian & $\exp{(\frac{1}{2\sigma^2}\Vert x_i - x_j \Vert_2^2 )}$ \\
    Factorized Laplacian & $\exp{(\frac{1}{\sqrt{2}\sigma}\Vert x_i - x_j \Vert_1)}$ \\
    Elliptical Laplacian & $\exp{(\frac{1}{\sqrt{2}\sigma}\Vert x_i - x_j \Vert_2)}$
    \end{tabular}

\end{table}

\begin{table}[!h]
    \caption{Description of the architectures employed in DiME-GAN experiments}
    \label{tab:dimegan_architecture}
    \centering
    \begin{tabular}{l|l}
    \textbf{Discriminator} & \textbf{Generator} \\\hline
    Input: $32\times 32 \times 3$ & Input: $1\times 1 \times 100$ \\ 
    $4\times4$ conv, $64$ out Channel, stride $2$, padding $1$ & $4\times4$ convTrans, $512$ out Channel, stride $1$, padding $0$ \\
    leakyReLU($0.2$) & leakyReLU $0.2$ \\
    $4\times4$ conv, $128$ out Channel, stride $2$, padding $1$ & $4\times4$ convTrans, $256$ out Channel, stride $2$, padding $1$ \\
    leakyReLU $0.2$  & leakyReLU $0.2$ \\
    $4\times4$ conv, $256$ out Channel, stride $2$, padding $1$ & $4\times4$ convTrans, $128$ out Channel, stride $2$, padding $1$ \\
    $4096$ fullyConnected, $64$ out Dims & leakyReLU $0.2$ \\
    ~ & $4\times4$ convTrans, $3$ out Channel, stride $2$, padding $1$ \\
    ~ & $\tanh$ \\
    \end{tabular}

\end{table}

\end{document}


%

%

\onecolumn
\aistatstitle{Instructions for Paper Submissions to AISTATS 2022: \\
Supplementary Materials}

\section{FORMATTING INSTRUCTIONS}

To prepare a supplementary pdf file, we ask the authors to use \texttt{aistats2022.sty} as a style file and to follow the same formatting instructions as in the main paper.
The only difference is that the supplementary material must be in a \emph{single-column} format.
You can use \texttt{supplement.tex} in our starter pack as a starting point, or append the supplementary content to the main paper and split the final PDF into two separate files.

Note that reviewers are under no obligation to examine your supplementary material.

\section{MISSING PROOFS}

The supplementary materials may contain detailed proofs of the results that are missing in the main paper.

\subsection{Proof of Lemma 3}

\textit{In this section, we present the detailed proof of Lemma 3 and then [ ... ]}

\section{ADDITIONAL EXPERIMENTS}

If you have additional experimental results, you may include them in the supplementary materials.

\subsection{The Effect of Regularization Parameter}

\textit{Our algorithm depends on the regularization parameter $\lambda$. Figure 1 below illustrates the effect of this parameter on the performance of our algorithm. As we can see, [ ... ]}

\vfill